\documentclass{article} 
\usepackage{iclr2027_arxiv,times}

\usepackage{amsmath,amsfonts,bm}

\def\eqref#1{equation~\ref{#1}}

\def\plaineqref#1{\ref{#1}}

\def\1{\bm{1}}

\DeclareMathAlphabet{\mathsfit}{\encodingdefault}{\sfdefault}{m}{sl}
\SetMathAlphabet{\mathsfit}{bold}{\encodingdefault}{\sfdefault}{bx}{n}

\usepackage{hyperref}
\usepackage[capitalize]{cleveref}
\usepackage{url}

\usepackage{amsmath,amssymb,mathtools} 
\usepackage{bbm}
\usepackage{booktabs,multirow,array}
\usepackage[table]{xcolor}
\usepackage{graphicx}
\usepackage{float}
\usepackage{pgfplots}
\usepackage{tikz}
\usetikzlibrary{positioning,arrows.meta}
\pgfplotsset{compat=1.18}
\usepackage{siunitx}
\usepackage{xspace}
\usepackage{etoc}

\title{GameWAM: A World Action Model for Video Games}

\author{Yuncheng Guo$^{1}$, Zhanqiu Zhang$^{2,\dagger}$, Yiwen Guo$^{3,\dagger}$, Weijia Li$^{4,\dagger}$ \\
$^{1}$Fudan University \\
$^{2}$LIGHTSPEED \\
$^{3}$Independent Researcher \\
$^{4}$Tsinghua Shenzhen International Graduate School \\
$^{\dagger}$Corresponding authors.
}

\iclrfinalcopy 
\begin{document}

\maketitle

\etocdepthtag.toc{main}

\begin{abstract}
Modern video games combine first-person perception, rapid visual changes,
persistent world state, and heterogeneous native controls. Existing game agents
map visual and task context directly to actions but lack explicit world
dynamics modeling, whereas interactive game world models predict visual futures from
supplied actions but do not serve as task policies. World Action Models (WAMs)
unify these objectives, but remain largely unexplored under the dynamics and
open-ended interaction of video games. We introduce
GameWAM, to our knowledge the first WAM for native
closed-loop gameplay and GUI control. GameWAM jointly generates future visual
observations and executable keyboard--mouse trajectories through parallel visual
and action generative processes with block-causal conditioning and flow
matching. To support joint world--action learning, we construct synchronized
gameplay and GUI trajectories. To handle heterogeneous native control, GameWAM predicts a gameplay/GUI mode per action step and
generates actions with mode-specific prediction distributions and
continuous-action normalization. For long-horizon interaction, block-cycle control coordinates prediction,
execution, and temporal context: it predicts beyond the committed horizon,
executes short action blocks, replans from new observations, and hierarchically
structures context from fine-grained within-cycle history to persistent
cross-cycle history. Experiments show competitive task success with fewer executed
native actions than the compared agents. We further uncover
\emph{Low-Frequency Action Source Imprinting} (LASI), in which low-frequency
components of the sampled action source systematically steer coarse generated
camera motion under fixed conditioning, revealing a source-sensitivity failure mode
in generative control. 
Project page is available at
\url{https://yunncheng.github.io/GameWAM/}.
\end{abstract}

\section{Introduction}

Modern video games demand pixel-based native control over rapidly changing
first-person observations, keyboard--mouse inputs, and the consequences of prior
actions. Existing solutions span behavioral pretraining, vision-language-action
models, hierarchical agent architectures, and large-scale interaction learning
~\citep{baker2022vpt,lifshitz2023steve1,wang2024jarvis,
wang2024omnijarvis,wang2026openha,wang2025gametars}.
Despite different system designs, their control ultimately maps visual and task
context to native or abstract actions, often through discretized, semantic, or
temporally compressed representations. While effective for behavior learning,
these objectives do not explicitly model how the visual world evolves under
executed actions, and their action abstractions may weaken the fine temporal and
metric structure needed for concurrent key control and continuous camera motion.
Interactive game world models provide the complementary capability: they learn
action-conditioned visual dynamics, but rely on actions supplied by a player or
external controller rather than selecting task-directed behavior themselves
~\citep{bruce2024genie,alonso2024diamond,che2025gamegenx,
guo2025mineworld,zhang2025matrixgame}. Control models therefore choose behavior
without explicitly modeling its visual consequences, while interactive world
models predict those consequences without serving as task policies. This
motivates a unified model of task-directed actions and their visual consequences.

\begin{figure*}[t]
    \centering
    \includegraphics[width=\textwidth]
    {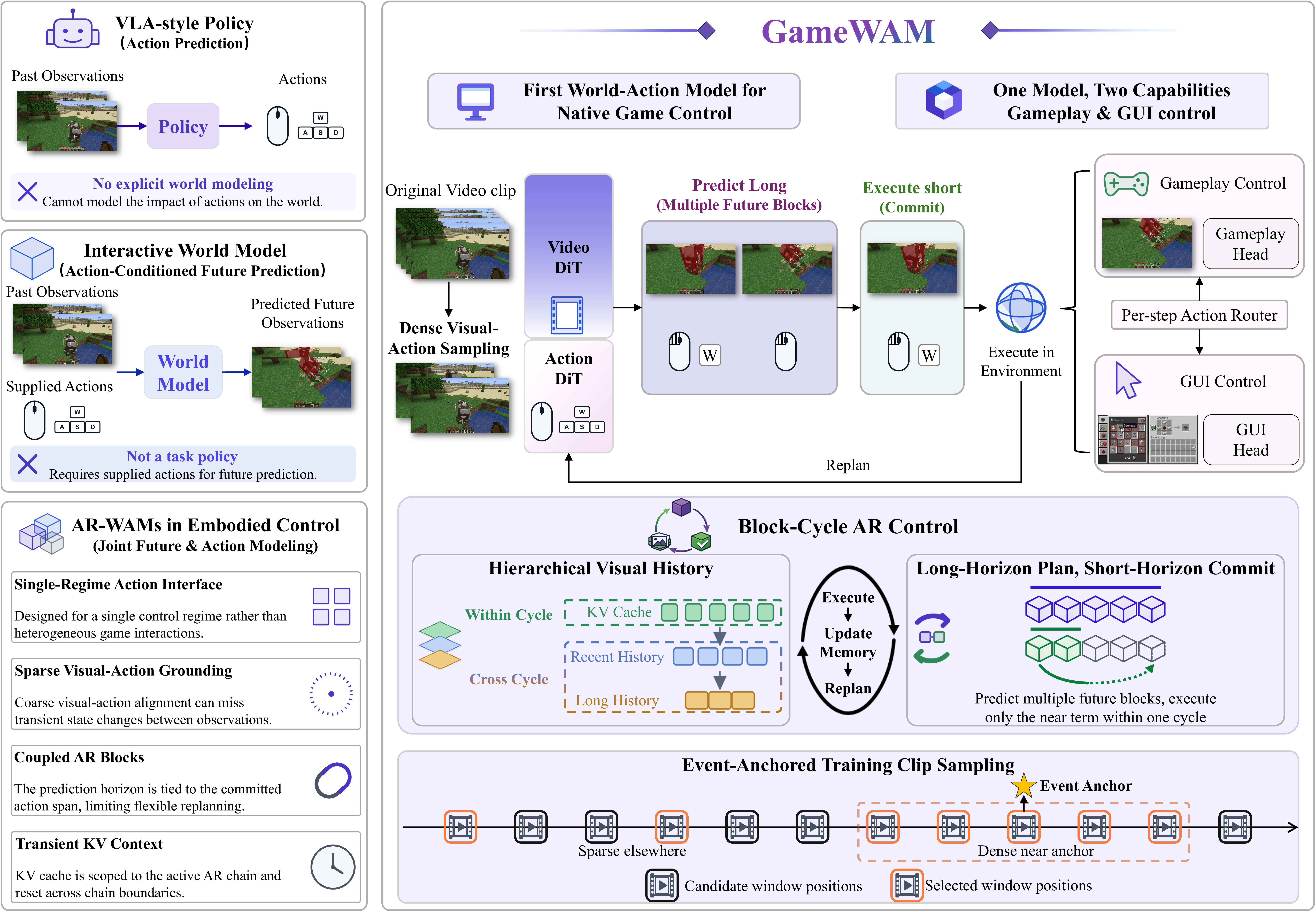}
    \vspace{-0.4cm}
    \caption{
    \textbf{Overview of GameWAM.}
    GameWAM jointly models future visual observations and native actions for closed-loop gameplay and GUI control.
    }
    \label{fig:gamewam_overview}
\vspace{-0.4cm}
\end{figure*}

World Action Models (WAMs) jointly generate future visual observations and
executable action trajectories, using visual prediction to provide dynamics-aware supervision for control. Recent WAMs
have shown promising results mainly in tabletop or bounded indoor robot
manipulation~\citep{ye2026dreamzero,yuan2026fastwam}, while their behavior under
rapidly changing first-person observations, persistent world state, and
repeated closed-loop interaction remains less understood. Modern video games
provide a scalable and controllable testbed for this regime: ego-motion,
dynamic entities, heterogeneous interaction modes, and persistent state create
diverse and abrupt visual changes, while environments remain precisely
instrumented, reliably reset, and repeatable under controlled
interventions~\citep{fan2022minedojo,raad2024sima}. Although games do not
reproduce every aspect of physical embodiment, they expose control models to
rapidly evolving interaction at a scale difficult to obtain in current
physical testbeds~\citep{khazatsky2024droid,pumacay2024colosseum}. We therefore
introduce GameWAM, to our knowledge the first WAM for native closed-loop
gameplay and GUI control. GameWAM jointly generates future visual observations
and executable keyboard--mouse trajectories from visual, instruction, and
other condition signals.

Realizing a WAM in games introduces challenges along action and temporal
dimensions. The action challenge is particularly prominent in native game
control: gameplay combines concurrent discrete keys, continuous camera motion,
and sparse mouse events, while GUI interaction reuses the same physical
channels for cursor motion, clicks, scrolling, and interface shortcuts. Thus,
identical action dimensions can have different semantics, scales, and
conditional distributions across regimes, and treating them as one
undifferentiated distribution can mix incompatible camera and cursor
statistics. GameWAM generates continuous and concurrent discrete dimensions within a
shared Action DiT flow, while a learned per-action router selects between
gameplay- and GUI-specific action-flow predictions at each action timestep.
The two regimes use separate normalization for their distinct control
distributions while retaining the same underlying physical controls.

The temporal challenge is amplified in rapidly changing interactive
environments by visual subsampling relative to native actions. Recent embodied
WAMs use temporal subsampling to control video-token cost, but coarse temporal
resolution can miss rapid ego-motion, transient targets, and brief interaction
events in games. Denser sampling captures these changes, yet under fixed token
and cache budgets the same resources cover less interaction time, shortening
both future look-ahead and retained past context. This trade-off especially affects chunk-wise autoregressive WAMs like
DreamZero~\citep{ye2026dreamzero}, where prediction, execution, and transient
KV accumulation proceed along a finite block chain. On the future side, denser
sampling shortens the real-time horizon covered by a fixed chain. Extending the
chain to recover longer look-ahead can also lengthen the committed action span
and delay feedback from updated observations. GameWAM instead decouples
prediction from commitment: each planning unit predicts $P$ actions but commits
only an $E$-step execution block, with $E<P$, before observing and replanning.
This predict-long/execute-short organization provides longer action look-ahead
while preserving frequent closed-loop feedback. During training, overlapping
$P$-step plans are anchored every $E$ steps, extending supervision beyond each
execution block while reusing the same $E$-spaced temporal organization rather
than lengthening the autoregressive chain. On the past side, denser sampling
similarly shortens the interaction span represented by a fixed-size K/V cache.
Retaining more blocks increases memory, while the transient cache is reset
when the finite chain ends, leaving subsequent chains without persistent access
to earlier executed observations. GameWAM therefore combines a bounded
within-cycle K/V cache with persistent visual history across cycle
boundaries: recent executed observations remain explicitly represented, while
older information is compressed into long-term history. This preserves
fine-grained current-cycle context alongside persistent multi-cycle
conditioning under bounded memory.

To support WAM training in this setting, we construct synchronized, WAM-ready
gameplay and GUI trajectories. GameWAM models their joint visual and control
dynamics with parallel Video and Action DiTs, block-causal attention, and joint
flow matching, while event-anchored sampling emphasizes event-rich segments.
Across Minecraft and ViZDoom, GameWAM achieves competitive closed-loop
performance; in Minecraft, it does so with fewer executed native actions than
the compared agents across all evaluated task categories. Our closed-loop evaluation further reveals an unexpected failure
mode: when the same sampled action source is reused across replanning steps,
some source realizations induce a persistent directional camera bias, including
repeated in-place rotation. Controlled frequency-domain interventions reveal that low-frequency components
of the sampled action source systematically steer coarse generated camera
motion under fixed conditioning. We term this phenomenon
\emph{Low-Frequency Action Source Imprinting} (LASI). LASI distinguishes
ordinary source-dependent sampling from a pathological low-frequency control
bias that can accumulate through repeated closed-loop interaction.
Our contributions are:
\begin{itemize}
    \item We introduce GameWAM, to our knowledge the first
    World Action Model for native closed-loop gameplay and GUI control.
    Built on parallel Video and Action DiTs, block-causal attention, and joint
    flow matching, GameWAM jointly generates future visual observations and
    executable keyboard--mouse action trajectories.

    \item We construct synchronized gameplay and GUI trajectories for joint
    world--action learning, combining standardized gameplay data with generated
    interface interactions.

    \item We develop a unified native-action and block--cycle formulation
    for heterogeneous, long-horizon game interaction. GameWAM supports gameplay
    and GUI control in a shared keyboard--mouse action space, while block--cycle control
    decouples long-horizon
    prediction from short-horizon execution and hierarchically organizes fine-grained within-cycle context and persistent cross-cycle history.
    Experiments demonstrate competitive closed-loop performance with fewer executed native
    actions than the compared agents.

    \item We uncover \emph{Low-Frequency Action Source Imprinting} (LASI), a
    source-sensitivity failure mode in which low-frequency components of the
    sampled action source steer coarse generated camera motion
    under fixed conditioning. Reuse across
replanning steps can accumulate into persistent directional bias in closed-loop
    interaction.
\end{itemize}

\begin{figure*}[t]
    \centering
    \includegraphics[width=\textwidth]
    {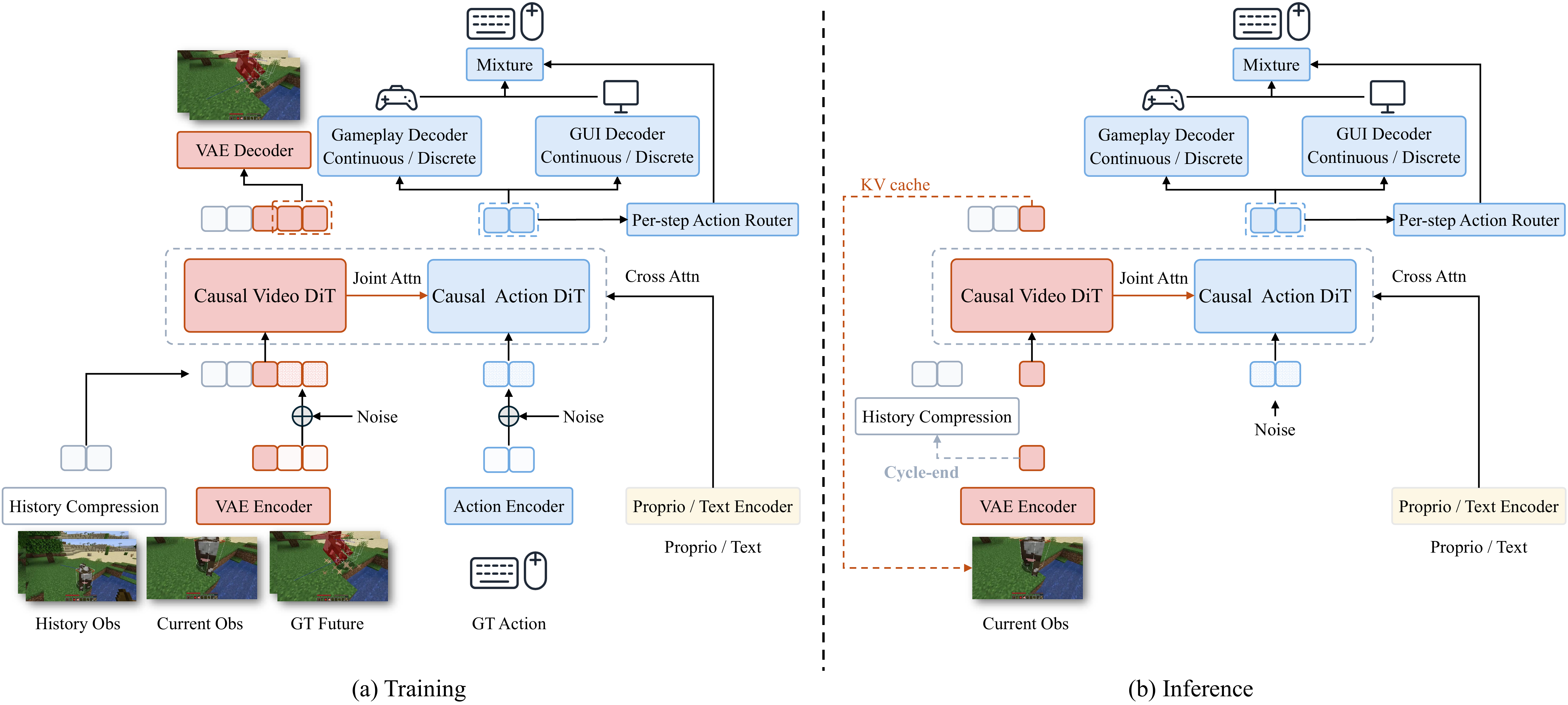}
    \vspace{-0.4cm}
    \caption{
    \textbf{GameWAM architecture for training and inference.}
    Parallel Video and Action DiTs jointly model future observations and native
    actions, with block-causal interaction and gameplay/GUI-specific action
    generation.
    }
    \label{fig:method_overview}
\vspace{-0.3cm}
\end{figure*}

\section{Related Work}

\subsection{Game Agents and Game World Models}

Video games provide scalable settings for native control and long-horizon
interaction. Earlier work established large-scale game environments,
demonstrations, behavioral learning, and open-ended evaluation
~\citep{guss2019minerl,fan2022minedojo,baker2022vpt,lifshitz2023steve1,
MineStudio,zheng2025mcu,kempka2016vizdoom,petrenko2020samplefactory}.
Subsequent game agents increasingly emphasize multimodal reasoning, memory,
hierarchical control, and generalist pretraining
~\citep{wang2024voyager,wang2024jarvis,wang2024omnijarvis,raad2024sima}.
Recent work more directly addresses action representation and scalable native
control. OpenHA~\citep{wang2026openha} systematically studies action
abstractions and introduces Chain of Action, using abstract actions as
intermediate reasoning steps before executable control.
Game-TARS~\citep{wang2025gametars} scales a unified native keyboard--mouse
action space through large-scale continual pretraining across heterogeneous
game and computer-use domains, while Lumine~\citep{tan2025lumine} unifies
perception, reasoning, and real-time keyboard--mouse control for long-horizon
3D open-world interaction.

Game world models learn controllable visual dynamics of interactive
environments. This line spans latent-action generation, diffusion world
modeling, real-time neural game simulation, and increasingly open-world
action-conditioned generation
~\citep{bruce2024genie,alonso2024diamond,valevski2024gamengen,
che2025gamegenx,guo2025mineworld,zhang2025matrixgame}.
WHAM~\citep{kanervisto2025wham} further models gameplay video together with
human controller actions. These developments build on broader advances in video
diffusion and token-based video generation
~\citep{ho2022videodiffusion,voleti2022mcvd,harvey2022flexiblevideo,
villegas2023phenaki,yu2023magvit}.

\subsection{World Action Models}

World Action Models connect predictive world modeling with generative control.
Learned world models have progressed from latent planning and imagination to
scalable transformer and continuous-control formulations
~\citep{hafner2019planet,hafner2020dreamer,hafner2021dreamerv2,
schrittwieser2020muzero,micheli2023iris,zhang2023storm,
hansen2024tdmpc2,hafner2025dreamerv3}. Generative control has similarly
developed structured action prediction through chunking, diffusion, latent
actions, and large-scale vision--language--action learning
~\citep{zhao2023act,chi2023diffusionpolicy,lee2024vqbet,
brohan2023rt1,zitkovich2023rt2,oneill2024openx,ghosh2024octo}, while video-
and latent-action methods connect predicted trajectories to control
~\citep{du2023unipi,du2024vlp,wu2024gr1,ye2025lapa}. Diffusion Transformers
and flow-based objectives provide scalable generative backbones
~\citep{peebles2023dit,ma2024sit,lipman2023flowmatching}. Related
world--action formulations have also been explored in offline model-based
RL~\citep{cheng2025scaling}.

More recent generative WAMs directly couple visual prediction with executable
action generation and increasingly revisit inference-time execution.
DreamZero~\citep{ye2026dreamzero} jointly predicts future video and actions for
closed-loop control, Fast-WAM~\citep{yuan2026fastwam} retains video co-training
while removing future-video generation at test time, and
FFDC-WAM~\citep{wang2026trustimagination} adapts execution by verifying
predicted futures against realized observations. GameWAM studies this emerging
paradigm in native closed-loop gameplay and GUI control.

\section{Data Construction and Training Sampling}
\label{sec:data_construction}

Joint world--action learning requires aligned observations and native controls.
For Minecraft, we construct three complementary WAM-ready training streams.
Regular VPT trajectories~\citep{baker2022vpt} provide broad naturalistic
interaction coverage and long temporal context. From the same recordings, we
additionally construct an Event-Anchored VPT dataset by identifying
MineStudio-style interaction events~\citep{MineStudio} from state transitions
and forming instruction-aligned event-centered trajectories. We further
generate MineStudio-based scripted GUI trajectories to broaden
interface-interaction coverage. All streams are standardized on a common
interaction timeline into synchronized observations, state, and native
keyboard--mouse actions with a control-mode label. Dataset construction details
are provided in \cref{app:data_details}.

\noindent\textbf{Event-anchored training clip sampling.}
Dataset construction determines the event-centered trajectories available for
training, whereas clip sampling separately determines their temporal sampling
density. Long interaction trajectories contain highly uneven supervision density:
transitions near an annotated event often contain the visual and action changes
most relevant to the instruction, while distant portions may consist largely
of traversal, incidental camera motion, or other weakly related behavior.
Uniform sampling would therefore devote substantial training capacity to less
informative regions. Within the Event-Anchored VPT stream, we instead sample
training clips densely around the event anchor and more sparsely away from it,
increasing exposure to event-relevant transitions while retaining surrounding
interaction context. The sampling configuration is in
\cref{app:experimental_protocol}.

\section{Method}
\label{sec:method}

\begin{figure*}[t]
    \centering
    \includegraphics[width=\textwidth]
    {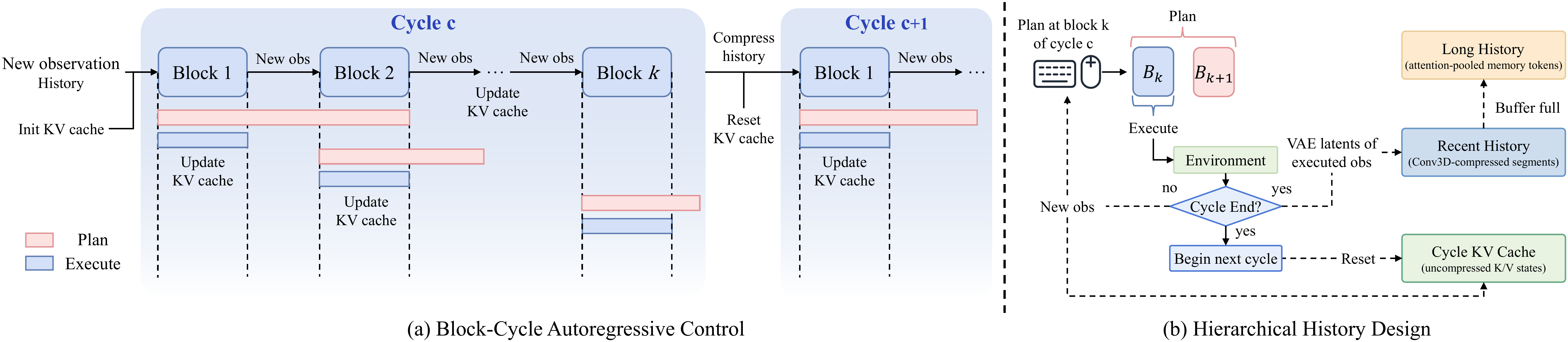}
    \vspace{-0.4cm}
    \caption{
    \textbf{Block--cycle control and hierarchical history.}
    GameWAM predicts beyond the committed horizon while executing only a short
    prefix, and combines a transient cycle-local K/V cache with persistent
    cross-cycle visual history.
    }
    \label{fig:gamewam_schedule}
\vspace{-0.3cm}
\end{figure*}

GameWAM is a block-causal world--action model for native gameplay and GUI
control. \cref{fig:gamewam_overview,fig:method_overview,fig:gamewam_schedule} summarize the model and rollout.

\subsection{Block-Causal World--Action Modeling}
\label{sec:block_causal_world_action}

At block $k$ of cycle $c$, define the conditioning context as
\begin{equation}
    \Gamma_{c,k}=(C_{c,k},\ell,H_c,s_{c,k}),
\label{eq:conditioning_context}
\end{equation}
where $C_{c,k}$ denotes the clean within-cycle observation context, $\ell$ the
language or task instruction, $H_c$ the persistent cross-cycle visual history,
and $s_{c,k}$ the optional proprioceptive state. The complete clean visual
prefix used by block-causal attention is
\begin{equation}
    \mathcal P_{c,k}=H_c\cup C_{c,k}.
\label{eq:clean_visual_prefix}
\end{equation}
A world--action block is $B_j=(\mathcal V_j,\mathcal A_j)$, with video latents $\mathcal V_j$ and
native actions $\mathcal A_j=[\mathcal A_j^{\mathrm{cont}};\mathcal A_j^{\mathrm{disc}}]$. For a plan spanning $R$ blocks, GameWAM models the block-causal joint distribution
\begin{equation}
p_\theta\!\left(B_{k:k+R-1}\mid\Gamma_{c,k}\right)
=
\prod_{r=0}^{R-1}
p_\theta\!\left(
B_{k+r}\mid\Gamma_{c,k},B_{k:k+r-1}
\right).
\label{eq:block_causal_factorization}
\end{equation}
The attention mask determines how the two modalities exchange information within
each conditional factor. We parameterize visual and action processes with
joint flow matching~\citep{lipman2023flowmatching} using parallel Video- and
Action DiT branches~\citep{peebles2023dit,ma2024sit}. Continuous and discrete
action coordinates remain components of one native action vector; discrete
ones are converted to executable binary decisions only after generation.
For modality $m\in\{v,a\}$,
\begin{equation}
X_{\sigma_m}^{m}=(1-\sigma_m)X_0^{m}+\sigma_m\epsilon^{m},
\qquad
U^{m}=\epsilon^{m}-X_0^{m}.
\label{eq:joint_flow}
\end{equation}
The parallel DiTs estimate $(\widehat U_\theta^v,\widehat U_\theta^a)$ from
the noisy modalities, their noise levels, and the causal context. At inference,
$X_1^m\sim\mathcal N(0,I)$ and the learned ODE is integrated from
$\sigma=1$ to $0$.

\noindent\textbf{Block-causal video--action masking.}
For target block $j$, let $\mathcal P_{c,j}$ denote its causally available
clean visual prefix, and let $\mathcal V_j$ and $\mathcal A_j$ denote the
noisy video and action variables of that block. We adopt a Fast-WAM-style
\citep{yuan2026fastwam} modality-decoupled mask, under which both modalities
share the same clean prefix while their noisy variables do not condition one
another:
\begin{equation}
\operatorname{Vis}(\mathcal A_j)
=
\mathcal P_{c,j}\cup\mathcal A_j,
\qquad
\operatorname{Vis}(\mathcal V_j)
=
\mathcal P_{c,j}\cup\mathcal V_j.
\label{eq:video_action_mask}
\end{equation}
The Video DiT encodes $\mathcal P_{c,j}$ into layer-wise K/V states shared by
both branches, so video and action supervision jointly shape the context used
for action generation. Instruction and proprioception remain available through
their corresponding conditioning paths. This preserves video
co-training while avoiding future-video denoising during action-only online
inference. We compare this default design with joint within-block
video--action attention in \cref{app:mask_comparison}.

\noindent\textbf{Per-action routing.}
Gameplay and GUI interaction share the same physical keyboard--mouse
coordinates but induce different conditional action distributions. GameWAM
therefore uses a timestep-wise router with gameplay- and GUI-specific
prediction branches. Let $r_\tau^*\in\{0,1\}$ denote the observed interaction
mode and $\rho_\tau$ the routing logit. At rollout, the predicted route
$\hat r_\tau$ selects the action-flow prediction for the entire native action
vector at timestep $\tau$:
\begin{equation}
\hat r_\tau=\mathbf 1\!\left[\operatorname{sigmoid}(\rho_\tau)>\tfrac12\right],
\qquad
\widehat U_\tau^{a}=(1-\hat r_\tau)\widehat U_\tau^{\mathrm{game}}
+\hat r_\tau\widehat U_\tau^{\mathrm{gui}}.
\label{eq:action_router}
\end{equation}
During training, $r_\tau^*$ selects the supervised prediction branch and also
provides the routing target; at rollout, the predicted route is used. Continuous
action coordinates are normalized with gameplay- or GUI-specific statistics,
whereas discrete coordinates retain the shared normalization. Validity masks
exclude controls unavailable in a given environment.

\subsection{Block-Cycle Control and Hierarchical History}
\label{sec:predict_execute}

As illustrated in \cref{fig:gamewam_schedule}, block--cycle control organizes
prediction, execution, and temporal context. A plan contains $P$ actions,
whereas only its first $E$ actions form the committed execution block:
\begin{equation}
\widehat{\mathbf A}_{c,k}^{\mathrm{plan}}\in\mathbb R^{P\times d_a},
\qquad E<P,\qquad
\widehat{\mathbf A}_{c,k}^{\mathrm{exec}}
=
\widehat{\mathbf A}_{c,k}^{\mathrm{plan}}[1{:}E].
\label{eq:block_cycle_plan}
\end{equation}
The plan spans $R=\lceil P/E\rceil$ execution intervals. After executing one
block, the agent receives a new observation and replans the overlapping
$P-E$ suffix. Only observations realized through executed interaction become
clean causal context; unexecuted look-ahead is discarded. During training, overlapping $P$-step plans anchored every $E$ steps are
formed from the same trajectory and teacher-forced in parallel under
block-causal visibility, avoiding sequential rollout of the additional
look-ahead.

GameWAM separates within-cycle realized context from persistent cross-cycle
history. Online, both form the clean prefix and its transient K/V cache;
persistent history comprises recent and long-term visual memory.

\noindent\textbf{Cycle K/V cache.}
During online rollout, $\mathcal P_{c,k}$ is materialized as layer-wise K/V
states. At cycle start, the cache is initialized from cross-cycle history and
current context; newly realized observations then extend only its current-cycle
portion. The cache is discarded at the cycle boundary and reinitialized from
$\mathcal P_{c+1,0}$; persistent information survives through $H_{c+1}$.

\noindent\textbf{Recent history.}
At a cycle boundary, the realized visual segment produced by executed actions is
encoded by the video VAE and compressed by a Conv3D tokenizer:
\begin{equation}
S_c=\operatorname{Conv3D}\!\left(\operatorname{VAE}(O_c^{\mathrm{exec}})\right),
\qquad
(R_{c+1},\bar S_c)=\operatorname{FIFO}_{K_R}(R_c,S_c).
\label{eq:recent_history_update}
\end{equation}
The buffer $R_c$ retains the latest $K_R$ segments; $\bar S_c$ denotes the
evicted segment when the buffer is full. Recent history therefore remains
explicitly addressable at a compressed spatial resolution.

\noindent\textbf{Long-term history.}
An evicted recent segment updates a fixed set of long-term memory slots through
attention pooling and a learned slot-wise gate:
\begin{equation}
\widetilde M_c=\operatorname{Attn}(M_c,\bar S_c,\bar S_c),
\qquad
M_{c+1}=\operatorname{LN}\!\left[M_c+g_c\odot(\widetilde M_c-M_c)\right].
\label{eq:long_history_update}
\end{equation}
If no segment is evicted, $M_{c+1}=M_c$. Learned slot and level embeddings
assign distinct temporal roles to different memory slots. The resulting
persistent history $H_c=[M_c;R_c]$ is retained across interaction-cycle
boundaries, unlike the transient cycle-local K/V state.

\subsection{Training Objective}
\label{sec:training_objective}

Training teacher-forces completed blocks with rollout block-causal visibility. The video flow loss is
\begin{equation}
\mathcal L_v
=
\mathbb E\!\left[
w_v(\sigma_v)\,\lVert\widehat U^v-U^v\rVert_2^2
\right].
\label{eq:video_flow_loss}
\end{equation}
For the selected interaction branch, the Action DiT jointly predicts the complete native
action vector. To prevent either continuous or discrete coordinates from
dominating, they are reduced separately. For
$g\in\{\mathrm{cont},\mathrm{disc}\}$,
\begin{equation}
\begin{gathered}
\mathcal L_a^{g}
=
\mathbb E\!\left[
w_a(\sigma_a)
\frac{\sum_{\tau,d\in\mathcal D_g}m_{\tau d}
(\widehat U_{\tau d}^{a}-U_{\tau d}^{a})^2}
{\sum_{\tau,d\in\mathcal D_g}m_{\tau d}}
\right],\quad g\in\{\mathrm{cont},\mathrm{disc}\};\\
\mathcal L_a
=
\lambda_{\mathrm{cont}}\mathcal L_a^{\mathrm{cont}}
+\lambda_{\mathrm{disc}}\mathcal L_a^{\mathrm{disc}}.
\end{gathered}
\label{eq:grouped_action_flow_loss}
\end{equation}
Here $m_{\tau d}$ masks padded or unavailable action coordinates, and
$\widehat U^a$ is selected by the observed interaction mode during training.
Routing is supervised via masked binary cross-entropy,
\begin{equation}
\mathcal L_{\mathrm{mode}}
=
\frac{\sum_\tau \nu_\tau\,
\operatorname{BCE}(\rho_\tau,r_\tau^*)}
{\sum_\tau \nu_\tau},
\label{eq:mode_loss}
\end{equation}
where $\nu_\tau$ indicates whether the interaction-mode label is valid. To
keep compressed history informative, we predict the current clean visual
feature $q_c$ from persistent history:
\begin{equation}
\mathcal L_{\mathrm{hist}}
=
D_{\mathrm{pred}}\!\left(
F(\operatorname{Pool}(H_c)),\operatorname{sg}(q_c)
\right),
\label{eq:history_loss}
\end{equation}
where $D_{\mathrm{pred}}$ combines cosine and squared-error distances. This
regularizes $H_c$ to retain cross-cycle predictive visual information, while
stop-gradient keeps the target fixed.
\begin{equation}
\mathcal L
=
\lambda_v\mathcal L_v
+\lambda_a\mathcal L_a
+\lambda_m\mathcal L_{\mathrm{mode}}
+\lambda_h\mathcal L_{\mathrm{hist}}.
\label{eq:total_objective}
\end{equation}

\section{Experiments}
\label{sec:experiments}

\subsection{Evaluation Setup}
\label{sec:benchmarks}

We evaluate GameWAM on two complementary closed-loop benchmarks:
Minecraft Universe (MCU) and the four-map ViZDoom suite. MCU evaluates long-horizon task completion under native control using task success and executed environment actions, jointly measuring control effectiveness and interaction efficiency. ViZDoom adds fast visual dynamics and
continuous interaction, evaluated by average episode
reward. Further details are provided in \cref{app:experimental_protocol}.

\subsection{Main Results on MCU}
\label{sec:mcu_results}

\begin{table*}[t]
    \centering
    \caption{
    \textbf{Evaluation results of Minecraft agents on the MCU benchmark with over
    800 tasks.}
    For a direct WAM-to-WAM comparison, we reproduce
    Fast-WAM~\citep{yuan2026fastwam} on the same game training data and MCU
    setup as GameWAM; results for all other baselines are taken from their
    original reports.
    For each category, we report average native interaction steps over successful
    episodes, ASR on the Mini subset (ASR Mini), and ASR over all tasks (ASR All).
    The \textit{ASR Mini} and \textit{ASR All} average 10 and 5 runs per task,
    respectively.
    Game PT. denotes large-scale policy pretraining or continual pretraining on
    interaction data spanning many game environments; single-game training and
    generic foundation-model pretraining are excluded.
    Results highlighted in \textcolor{red}{red} and \textcolor{blue}{blue}
    denote the best and second-best performance, respectively.
    }
    \label{tab:mcu}

    \resizebox{\textwidth}{!}{
    \renewcommand\arraystretch{1.2}
    \begin{tabular}{@{}lccccccccccccc c@{}}

\toprule
 &  & \multicolumn{3}{c}{Embodied Tasks} &  &
 \multicolumn{3}{c}{GUI Tasks} &  &
 \multicolumn{3}{c}{Combat Tasks} &
 \multicolumn{2}{c}{Avg} \\

\cmidrule(lr){3-5}
\cmidrule(lr){7-9}
\cmidrule(lr){11-13}
\cmidrule(l){14-15}

\multirow{-2}{*}{Model} &
\multirow{-2}{*}{\begin{tabular}[c]{@{}l@{}}Game PT.\end{tabular}} &
Steps $\downarrow$ & ASR Mini $\uparrow$ & ASR All $\uparrow$ &  &
Steps $\downarrow$ & ASR Mini $\uparrow$ & ASR All $\uparrow$ &  &
Steps $\downarrow$ & ASR Mini $\uparrow$ & ASR All $\uparrow$ &
Mini $\uparrow$ & All $\uparrow$ \\

\midrule

\multicolumn{15}{l}{
\cellcolor[HTML]{ECF4FF}{
\textit{\textbf{Instruction-Conditioned Policies}}
}} \\

VPT & $\times$
& 377 & $10.1^{\pm3.6}$ & $6.0^{\pm11.4}$ &
& 398 & $0.7^{\pm0.1}$ & $0.8^{\pm3.3}$ &
& 396 & $3.6^{\pm7.7}$ & $3.6^{\pm7.7}$ &
4.8 & 3.5 \\

STEVE-1 & $\times$
& 384 & $8.4^{\pm3.0}$ & $8.0^{\pm17.0}$ &
& 391 & $0.0$ & $3.2^{\pm8.4}$ &
& 395 & $4.9^{\pm1.8}$ & $3.9^{\pm12.0}$ &
4.4 & 5.0 \\

ROCKET-1 & $\times$
& 392 & $19.2^{\pm6.1}$ & $18.9^{\pm24.3}$ &
& -- & $0.0$ & $0.0$ &
& 320 & $29.8^{\pm9.0}$ & $27.9^{\pm29.3}$ &
16.3 & 15.6 \\

JARVIS-VLA & $\times$
& 305 & $31.0^{\pm12.7}$ & $30.0^{\pm35.4}$ &
& 339 & $25.3^{\pm5.7}$ & $25.1^{\pm23.9}$ &
& 352 & $18.3^{\pm5.2}$ & $18.5^{\pm22.7}$ &
24.9 & 24.5 \\

\midrule

\multicolumn{15}{l}{
\cellcolor[HTML]{F0FBEF}{
\textit{\textbf{VLM-based Agents}}
}} \\

LatentHA & $\times$
& 363 & $27.3^{\pm37.4}$ & $24.4^{\pm31.1}$ &
& 393 & $3.5^{\pm8.7}$ & $3.0^{\pm7.5}$ &
& 371 & $8.2^{\pm15.6}$ & $8.5^{\pm17.9}$ &
13.0 & 12.0 \\

MotionHA & $\times$
& 336 & $31.6^{\pm10.1}$ & $27.4^{\pm35.2}$ &
& -- & $0.0$ & $0.0$ &
& 392 & $9.1^{\pm3.9}$ & $4.3^{\pm10.8}$ &
13.6 & 10.6 \\

GroundingHA & $\times$
& 290 & $39.7^{\pm13.7}$ & $37.1^{\pm38.5}$ &
& 380 & $3.7^{\pm2.3}$ & $6.7^{\pm10.8}$ &
& 346 & $28.2^{\pm6.2}$ & $26.5^{\pm23.4}$ &
23.9 & 23.4 \\

SkillHA & $\times$
& 365 & $13.8^{\pm7.7}$ & $11.3^{\pm14.5}$ &
& 397 & $3.4^{\pm0.8}$ & $6.3^{\pm9.2}$ &
& 393 & $3.4^{\pm0.8}$ & $6.5^{\pm9.3}$ &
6.9 & 8.0 \\

TextVLA & $\times$
& 321 & $23.9^{\pm8.9}$ & $27.0^{\pm17.0}$ &
& 291 & $14.0^{\pm4.1}$ & $25.8^{\pm14.3}$ &
& 317 & $27.1^{\pm11.8}$ & $10.0^{\pm6.1}$ &
21.7 & 20.9 \\

OpenHA & $\times$
& 287 & $37.0^{\pm15.9}$ & $30.1^{\pm13.9}$ &
& 314
& \textcolor{blue}{$33.3^{\pm13.3}$}
& \textcolor{blue}{$32.5^{\pm9.2}$} &
& 316
& \textcolor{red}{$40.0^{\pm19.6}$}
& \textcolor{blue}{$31.9^{\pm13.7}$} &
36.8 & \textcolor{blue}{31.5} \\

\textcolor{black!40}{Game-TARS}
& \textcolor{black!40}{$\checkmark$}
& \textcolor{black!40}{373}
& --
& \textcolor{black!40}{$50.4^{\pm20.7}$} &
& \textcolor{black!40}{406}
& --
& \textcolor{black!40}{$39.1^{\pm27.5}$} &
& \textcolor{black!40}{372}
& --
& \textcolor{black!40}{$38.1^{\pm24.6}$} &
--
& \textcolor{black!40}{42.5} \\

\midrule

\multicolumn{15}{l}{
\cellcolor[HTML]{F4F0FF}{
\textit{\textbf{World Action Models}}
}} \\

Fast-WAM & $\times$
& \textcolor{blue}{165}
& \textcolor{blue}{$66.0^{\pm25.4}$}
& \textcolor{blue}{$41.2^{\pm36.6}$} &
& \textcolor{blue}{234}
& $23.0^{\pm19.5}$
& $18.4^{\pm22.1}$ &
& \textcolor{blue}{261}
& $31.0^{\pm29.1}$
& $16.3^{\pm22.4}$ &
\textcolor{blue}{40.0}
& 25.2 \\

\textbf{GameWAM} & $\times$
& \textcolor{red}{138}
& \textcolor{red}{$70.0^{\pm32.2}$}
& \textcolor{red}{$47.5^{\pm36.0}$} &
& \textcolor{red}{155}
& \textcolor{red}{$43.0^{\pm32.9}$}
& \textcolor{red}{$60.0^{\pm38.6}$} &
& \textcolor{red}{203}
& \textcolor{blue}{$39.0^{\pm25.9}$}
& \textcolor{red}{$32.2^{\pm30.2}$} &
\textcolor{red}{50.7}
& \textcolor{red}{46.6} \\

\bottomrule
\end{tabular}
}
\end{table*}

\cref{tab:mcu} shows that GameWAM achieves the highest average success rates (ASR) on both the Mini and full task sets while requiring substantially fewer steps per successful episode across all task categories. The interaction-efficiency gains are particularly pronounced on embodied and GUI tasks.

\subsection{Main Results on ViZDoom}
\label{sec:vizdoom_results}

\begin{figure*}[t]
    \centering
    \includegraphics[width=\textwidth]
    {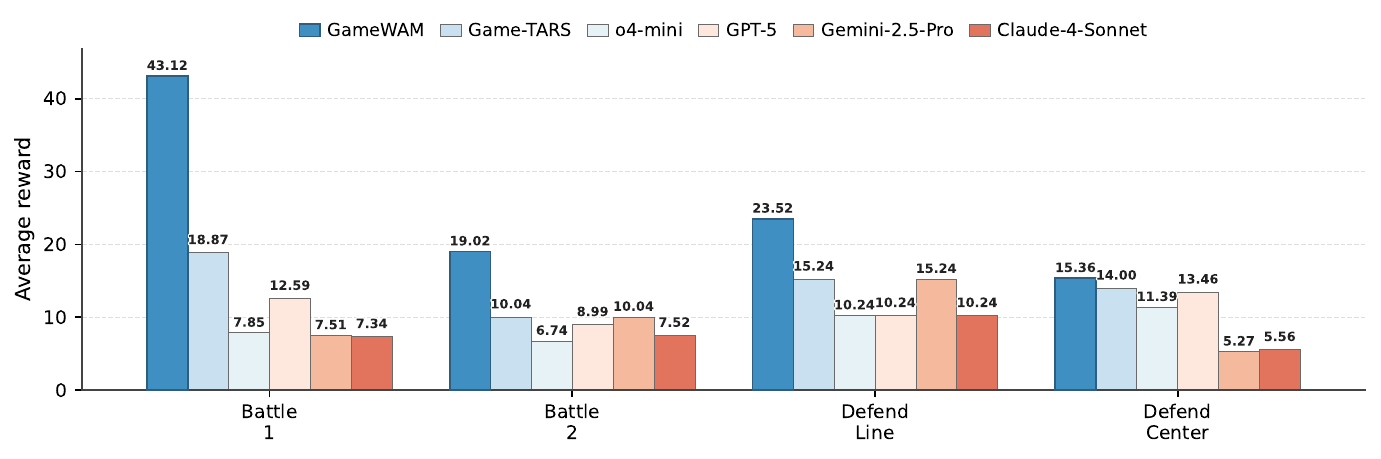}
    \vspace{-0.4cm}
    \caption{\textbf{Evaluation results on the four-map ViZDoom FPS benchmark.} The reported average reward for each map is calculated over 50 episodes.}
    \label{fig:vizdoom}
\vspace{-0.3cm}
\end{figure*}

\cref{fig:vizdoom} evaluates GameWAM under ViZDoom's distinct dynamics
and native control. GameWAM consistently improves over Game-TARS and achieves
competitive or leading average rewards relative to the compared multimodal
agents across all four scenarios. Together with the MCU results, this
demonstrates the formulation's effectiveness across diverse game environments.

\subsection{Ablation Analysis}
\label{sec:ablation}

\cref{tab:model_ablation} shows that removing future-video supervision
or using coarser temporal sampling produces the largest performance
degradation. Event-anchored clip sampling also contributes substantially,
consistent with emphasizing behaviorally informative transitions during
training. A unified gameplay/GUI action distribution underperforms the
mode-specific formulation, supporting separate prediction and normalization
for heterogeneous native controls. Matching the prediction and execution
horizons ($P=E$) further degrades ASR, indicating a benefit from predicting
beyond the committed horizon while retaining frequent feedback. Cross-cycle
history has a smaller and task-dependent effect: GUI and combat performance
declines without it, whereas the embodied point estimate increases slightly,
suggesting greater value for tasks that rely on information across replanning
cycles.

\begin{table}[t]
\centering
\small
\caption{
\textbf{GameWAM ablations on MCU Mini.}
}
\label{tab:model_ablation}
\setlength{\tabcolsep}{5.0pt}
\renewcommand{\arraystretch}{1.08}
\begin{tabular}{@{}lcccc@{}}
\toprule
Variant & Embodied & GUI & Combat & Avg \\
\midrule

Full GameWAM
    & $70.0^{\pm32.2}$
    & $\mathbf{43.0^{\pm32.9}}$
    & $\mathbf{39.0^{\pm25.9}}$
    & $\mathbf{50.7}$ \\

Action-only supervision
    & $60.0^{\pm27.6}$
    & $34.0^{\pm35.8}$
    & $13.0^{\pm13.5}$
    & 35.7 \\

Coarser temporal sampling
    & $60.0^{\pm29.7}$
    & $34.0^{\pm40.0}$
    & $16.0^{\pm36.7}$
    & 36.7 \\

No event-anchored clip sampling
    & $64.0^{\pm28.7}$
    & $33.0^{\pm30.0}$
    & $17.0^{\pm14.9}$
    & 38.0 \\

Unified action distribution
    & $59.0^{\pm25.5}$
    & $35.0^{\pm27.7}$
    & $21.0^{\pm29.1}$
    & 38.3 \\

Matched prediction--execution horizon ($P=E$)
    & $63.0^{\pm33.8}$
    & $32.0^{\pm36.8}$
    & $29.0^{\pm28.1}$
    & 41.3 \\

No cross-cycle history
    & $\mathbf{75.0^{\pm30.1}}$
    & $31.0^{\pm29.8}$
    & $34.0^{\pm31.0}$
    & 46.7 \\

\bottomrule
\end{tabular}
\end{table}

\subsection{Low-Frequency Action Source Imprinting}
\label{sec:lasi}

\begin{figure*}[t]
    \centering
    \includegraphics[width=\textwidth]
    {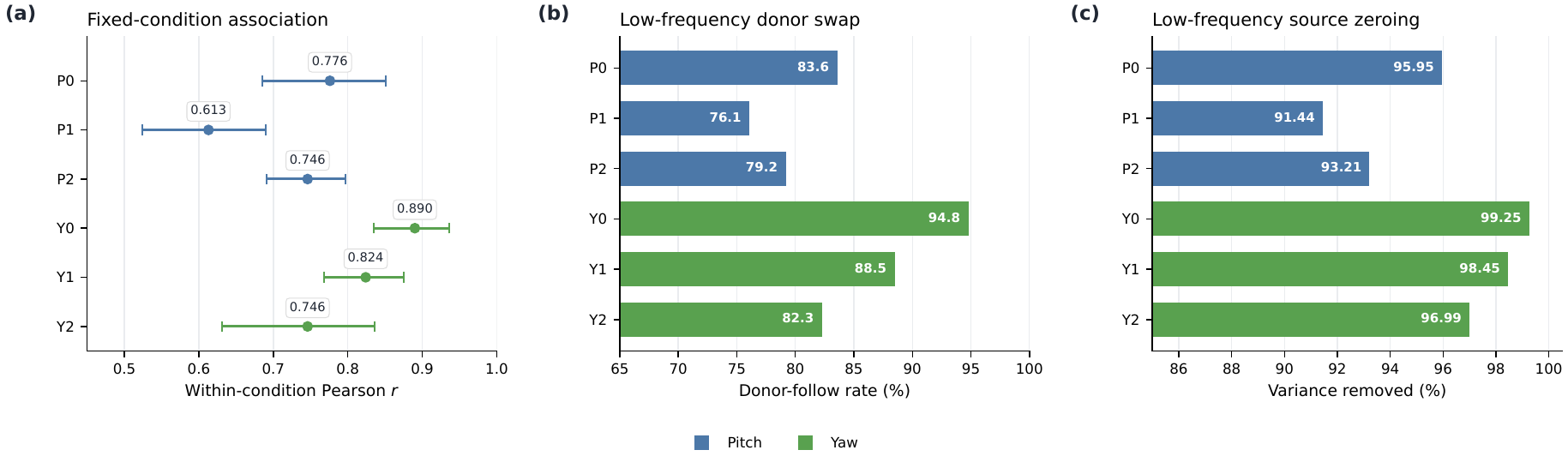}
    \vspace{-0.4cm}
    \caption{
    \textbf{Controlled evidence for LASI.}
    Fixed-condition association, low-frequency source replacement, and source
    zeroing quantify the influence of sampled low-frequency components on
    generated camera motion.
    }
    \label{fig:lasi_causal}
\vspace{-0.4cm}
\end{figure*}

Our investigation of action-source sensitivity arose from a closed-loop failure
observed during evaluation. Reusing the same sampled source across replanning
cycles caused some realizations to induce persistent directional camera bias
and repeated in-place rotation, in severe cases leaving almost no task
completion. Resampling between cycles largely removed this pattern, motivating
a controlled study of how source variation shapes generated actions. Source dependence is expected in diffusion and flow-based generation because
the sampled source initializes the generative trajectory and selects among
possible outputs. LASI is more structured: under fixed
conditioning, low-temporal-frequency source components coherently steer coarse
camera motion. We therefore fix the visual, historical, and proprioceptive
context. Let $X_0,Z\in\mathbb{R}^{H_a\times D_a}$ denote a generated
continuous-action chunk and its sampled source. We apply an orthonormal DCT
along the action horizon,
\begin{equation}
    \widetilde{X}_0=CX_0,\qquad
    \widetilde{Z}=CZ,\qquad
    C^\top C=I ,
\end{equation}
where modes $0$--$2$ form the low-frequency band.

\cref{fig:lasi_causal} provides three complementary tests. Under fixed
conditioning, matching low-frequency source and action coefficients remain
strongly associated, with yaw DCT0 reaching $r=0.890$. Replacing only source
modes $0$--$2$ makes the corresponding yaw DCT0 output follow the donor in
$94.8\%$ of trials, while zeroing the same band removes $99.25\%$ of the
associated output variance. Together, these interventions establish causal
evidence that low-frequency action-source components steer coarse generated camera motion rather than a correlation induced by changing conditions.

LASI is therefore distinct from ordinary sampling diversity: a sampled source
can acquire a coherent low-frequency control bias that becomes harmful when
reused through repeated closed-loop interaction. In our closed-loop evaluation,
we consequently resample the action source at each replanning step instead of reusing a fixed source throughout the rollout. This mitigates coherent
episode-level accumulation without removing the underlying source sensitivity.
Frequency selectivity, single-forward transfer, model-path amplification, and
closed-loop alignment are in \cref{app:lasi}.

\section{Conclusion}
\label{sec:conclusion}

We introduced GameWAM, a World Action Model for native closed-loop gameplay
and GUI control that jointly models future visual observations and executable
keyboard--mouse actions. Across Minecraft and ViZDoom, GameWAM achieves
competitive closed-loop performance, while using fewer executed native actions
than the compared agents in Minecraft. We further identify
\emph{Low-Frequency Action Source Imprinting} (LASI), showing that sampled
low-frequency action-source structure can induce persistent control bias when
reused across replanning steps. Together, these results support joint
world--action modeling as a promising direction for interactive game control
while exposing a source-sensitivity failure mode specific to generative action
policies.

\section*{AI Use Statement}

Generative AI tools were used for language polishing and editorial assistance,
including improving grammar, clarity, concision, and academic phrasing, as well
as refining the presentation of author-produced experimental results. The core
research questions, method, model design, data construction, implementation,
experimental results, and numerical measurements were developed or produced by
the authors. All AI-assisted text was manually reviewed and checked against the
implementation, experimental records, and underlying results, and was revised
by the authors where necessary. The authors take full responsibility for the
final content of this work, including all claims, analyses, and artifacts.

\section*{Ethics Statement}

This work studies learned closed-loop control in simulated video-game
environments. We conduct no new human-subject experiments and collect no
sensitive personal data as part of this study. The reported experiments are
restricted to digital game environments and do not involve deployment in
physical systems. While GameWAM is developed for research on native game
control, models capable of operating general keyboard--mouse interfaces could
in principle be adapted to other interactive software; deployment beyond the
controlled environments studied here should therefore consider authorization,
safety, and potential misuse. 

\section*{Reproducibility Statement}

We provide detailed descriptions of the model formulation, training objectives,
and closed-loop control procedure in the main text and
\cref{app:extended_method}. \cref{app:data_details} documents
trajectory alignment, native action interfaces, training data construction,
and training-sequence semantics, while
\cref{app:experimental_protocol} reports the architecture, temporal and
history configurations, training-data mixture and clip-sampling settings,
optimization details, objective weights, sampling configuration, and
evaluation protocols. Additional quantitative and qualitative analyses are
provided in \cref{app:additional_analysis}.

We will publicly release the training and evaluation code, data-processing
scripts, constructed datasets, trained GameWAM model weights, and associated
configuration files. The released resources will include the complete
procedures required to reproduce the reported experiments, subject to
applicable third-party licensing and redistribution requirements. Together
with the detailed methodological descriptions and experimental protocols
provided in this paper, these resources are intended to facilitate
reproduction of our results and further research on world action models for
interactive game control.

\bibliography{iclr2027_conference}
\bibliographystyle{iclr2027_conference}

\clearpage

\appendix

\etocdepthtag.toc{appendix}

\begingroup
\etocsettagdepth{main}{none}
\etocsettagdepth{appendix}{subsection}
\etocsettocstyle
  {\section*{Appendix Contents}}
  {}
\tableofcontents
\endgroup

\clearpage

\section{Extended Method}
\label{app:extended_method}

The main text defines the GameWAM formulation, block-causal visibility,
prediction--execution decomposition, hierarchical history, and learning
objective. This section expands their computational realization under
teacher-forced training and online interaction. We first clarify how the same
block-causal information boundary is instantiated under the two execution
settings and how realized context is organized within and across interaction
cycles. We then detail the cross-cycle history update, transformer-level
multimodal computation and masking variants, timestep-wise gameplay/GUI action
routing, and the supervision semantics of the training objective. Concrete
architectural dimensions, rollout horizons, memory capacities, and training
configurations are reported separately in
\cref{app:experimental_protocol}.

\subsection{Teacher-Forced Training and Online Rollout}
\label{app:teacher_forced_rollout}

Teacher-forced training and online rollout share the same block-causal
information boundary but differ in how the clean causal prefix becomes
available. During training, complete trajectories are available from the
dataset. We therefore construct prediction anchors from the $E$-spaced execution-aligned
temporal grid and provide each anchor with its corresponding ground-truth
causal prefix. Future visual and action variables within the prediction horizon
remain generation targets and are corrupted according to the flow objective,
while observations beyond the prediction boundary remain masked. Teacher
forcing therefore changes the computational realization of the clean prefix,
rather than the conditional information available to the model.

This organization preserves the realized-versus-future separation used during
online interaction. Although multiple prediction anchors may be constructed
from the same trajectory, each anchor only receives observations that are
causally available at its own prediction boundary. Ground-truth observations
that occur later in the trajectory do not become additional context for earlier
anchors, and predicted future variables from one anchor are never promoted into
the clean prefix of another anchor. The resulting training computation therefore preserves the causal semantics
of closed-loop execution while allowing supervision to be formed directly from
complete trajectories.

Online rollout realizes the same dependency structure sequentially. At each
decision point, GameWAM predicts a $P$-step action plan but commits only the
first $E$ actions. The environment is then advanced by this committed
execution block, producing a new realized observation for the next decision.
The model subsequently replans from the updated causal context, while the
unexecuted suffix of the previous prediction is discarded. Consequently, only
observations obtained through executed interaction can enter the clean prefix
or persistent history.

The prediction horizon and execution horizon play different roles in this
process. The block--cycle temporal organization is defined by the execution
horizon $E$: prediction anchors, committed blocks, and interaction updates are
aligned to the same $E$-spaced temporal grid. In contrast, the prediction
horizon $P$ determines how far into the future each anchor is supervised. For
the $j$-th prediction anchor,
\begin{equation}
    a_j=jE,
    \qquad
    Y_j=A[a_j+1:a_j+P],
    \label{eq:overlapping_plan_targets}
\end{equation}
where $Y_j$ denotes the corresponding future action target. When $P>E$,
neighboring prediction targets overlap by $P-E$ actions. Increasing $P$
therefore extends the predictive supervision attached to each
execution-aligned anchor without changing the underlying block or cycle
boundaries.

This execution-aligned overlapping supervision increases the predictive
coverage extracted from each interaction trajectory. Because the
execution-aligned anchors remain fixed by $E$, increasing $P$ extends the
forecast range associated with each anchor rather than requiring additional
execution-aligned samples. Realized transitions in overlapping regions can
therefore be reused across multiple forecast offsets and supervised from
different causally valid prefixes while maintaining the same decision cadence.
For example, a transition that is farther in the future for one anchor may
become a near-term prediction target for the next anchor. GameWAM therefore
combines frequent $E$-spaced decision points with longer $P$-step predictive
supervision, rather than requiring the training sequence to adopt a longer
execution interval.

Because complete trajectories are available during training, these overlapping
targets do not require autoregressive rollout of intermediate predicted
blocks. Each prediction anchor is supervised using its own ground-truth causal
prefix, avoiding the need to generate previous blocks merely to construct later
conditioning context. This distinction does not make larger prediction
horizons computationally free: increasing $P$ still introduces additional
future targets and associated prediction losses. Instead, the benefit of the
organization is that longer-horizon supervision and frequent execution-aligned
anchors coexist without extending the autoregressive commitment chain.

During action-only online evaluation, iterative denoising of future visual
variables is omitted. The realized visual prefix remains available as the
context for action generation, while only the unobserved future video
generation process is removed. The resulting computation preserves the same
closed-loop execution semantics while removing future-video generation that is not required for immediate
control.

\subsection{Temporal Context Within and Across Interaction Cycles}
\label{app:temporal_context}

The preceding training and rollout schedules organize temporal context at
multiple timescales. The cross-cycle state $H_c$ is already defined in the main text and is
the historical representation that survives an interaction-cycle boundary.
Within an active cycle, we additionally denote by $C_{c,j}$ the realized visual
context available before decision $j$. It contains current-cycle observations
that have actually been reached through executed actions. As defined in the main
text, the complete clean visual prefix combines persistent cross-cycle history
with the realized context accumulated within the active cycle:
\begin{equation}
\mathcal P_{c,j}=H_c\cup C_{c,j}.
\label{eq:app_clean_visual_prefix}
\end{equation}
This decomposition is informational rather than architectural: $H_c$ specifies
what remains accessible from earlier cycles, whereas $C_{c,j}$ preserves the
fine-grained context accumulated inside the current cycle. Predicted but
unexecuted future variables belong to neither term.

Under teacher forcing, $C_{c,j}$ is obtained directly from the corresponding
ground-truth prefix of the sampled trajectory. During online rollout it can
only grow after the environment has returned a new realized observation,
\begin{equation}
C^{\mathrm{inf}}_{c,j+1}
=
C^{\mathrm{inf}}_{c,j}\cup\{o^{\mathrm{env}}_{c,j+1}\}.
\label{eq:app_online_context_update}
\end{equation}
Hence a predicted future observation is not treated as history merely because
it has been generated; only the observation reached after execution can extend
the within-cycle context.

The online K/V cache is a computational representation of the instantiated
clean prefix $\mathcal P_{c,j}$, rather than a third source of temporal context.
At the beginning of a cycle, the persistent cross-cycle history $H_c$ and the
initial realized within-cycle context are processed together to form the clean
layer-wise K/V states. As interaction continues, only newly realized
observations extend the current-cycle portion of this cache. Denoting the
retained clean-prefix representation by $\mathcal Z^{\mathrm{KV}}_{c,j}$,
\begin{equation}
\mathcal Z^{\mathrm{KV}}_{c,j}
=
\Psi_\theta\!\left(\mathcal P^{\mathrm{inf}}_{c,j}\right)
=
\Psi_\theta\!\left(
H_c\cup C^{\mathrm{inf}}_{c,j}
\right),
\label{eq:app_kv_realization}
\end{equation}
where $\Psi_\theta$ denotes the layer-wise transformation of the clean prefix.
During teacher-forced training, the analogous clean representations are formed
transiently inside the parallel forward computation. During online inference,
they are retained and incrementally extended as new observations are realized.
At a cycle boundary the transient K/V state is discarded, while the completed
interaction updates the persistent history used to construct
$\mathcal P_{c+1,0}$ for the next cycle.

Thus, $H_c$ is the persistent state across cycles, $C_{c,j}$ is the
fine-grained realized context accumulated within the active cycle, and
$\mathcal Z^{\mathrm{KV}}_{c,j}$ is their transient processed representation
during online computation. Training and inference differ in the computational
lifetime of this representation, not in the semantic source of historical
information.

\subsection{Cross-Cycle Hierarchical Memory}
\label{app:hierarchical_history}

The main text defines the persistent history as
$H_c=[M_c;R_c]$, where $R_c$ retains recent executed segments and $M_c$
summarizes older segments evicted from the recent buffer. Here we specify the
temporal encoding and long-term update used to construct these two components.
As in Eq.~\ref{eq:recent_history_update}, the representation
\begin{equation}
S_c
=
\operatorname{Conv3D}
\left(
\operatorname{VAE}(O_c^{\mathrm{exec}})
\right)
\label{eq:app_history_segment}
\end{equation}
is formed only from observations realized through execution. The Conv3D
tokenizer reduces spatial resolution while preserving the temporal ordering
within the segment.

\paragraph{Temporally encoded segment tokens.}
Both recent-history readout and long-term memory updates operate on the same
compressed segment representation $S_c$. Let $s_{c,\tau,k}$ denote the token at
temporal position $\tau$ and compressed spatial position $k$. Each token is
augmented with embeddings for its within-segment temporal position and spatial
token identity, together with a relative-time embedding. For a segment spanning
raw temporal indices $[b,e]$ and a reference time $t$, we use
\begin{equation}
\eta_c(t)
=
\left[
\log\!\left(1+\frac{t-b}{T_{\mathrm{cyc}}}\right),\;
\log\!\left(1+\frac{(t-e)_+}{T_{\mathrm{cyc}}}\right),\;
\frac{e-b}{T_{\mathrm{cyc}}}
\right],
\label{eq:history_time_descriptor}
\end{equation}
where $T_{\mathrm{cyc}}$ is the cycle stride in the native interaction
timeline. A small embedding network $\psi$ maps this descriptor to the model
dimension. The resulting token representation has the form
\begin{equation}
u_{c,\tau,k}(t)
=
s_{c,\tau,k}
+
e^{\mathrm{time}}_{\tau}
+
e^{\mathrm{space}}_{k}
+
\psi\!\left(\eta_c(t)\right).
\label{eq:history_token_encoding}
\end{equation}
Recent-history readout and long-term writing use separate role embeddings and
normalization parameters on top of $u_{c,\tau,k}$, while sharing the underlying
compressed segment tokens and temporal descriptor. For recent segments, $t$
is the current interaction time, so their relative-age encoding evolves as the
rollout proceeds. For a segment entering long-term memory, the reference is its
overflow time from the recent buffer.

\paragraph{Recent buffer and overflow.}
The recent branch retains the latest $K_R$ completed segments explicitly:
\begin{equation}
(R_{c+1},\bar S_c)
=
\operatorname{FIFO}_{K_R}(R_c,S_c).
\label{eq:app_recent_fifo}
\end{equation}
No long-term write occurs while the buffer has free capacity. Once it is full,
the oldest segment $\bar S_c$ is evicted and becomes the source of a long-term
memory update. Thus, recent observations remain individually addressable until
they leave the recent buffer; hierarchical summarization is triggered by
overflow rather than applied to every completed cycle.

\paragraph{Multi-timescale long-term update.}
Let
$M_c\in\mathbb R^{L\times K_M\times d}$ contain $K_M$ persistent slots at each
of $L$ temporal levels. The evicted segment is first converted to long-memory
source tokens $B_c$ using the long-term role encoding described above. Each
memory slot attends directly to the complete overflow segment:
\begin{equation}
\widetilde M_{\ell,k}
=
\operatorname{Attn}
\left(
M_{\ell,k}
+
e^{\mathrm{level}}_{\ell}
+
e^{\mathrm{slot}}_{k},
\;
B_c,\;
B_c
\right).
\label{eq:hierarchical_memory_candidate}
\end{equation}
The level and slot embeddings distinguish the roles of different persistent
states, while the update rate varies explicitly across temporal levels. Let
$h_\ell$ denote the characteristic half-life of level $\ell$ and let $\Delta$
denote the duration represented by the incoming segment. The corresponding
update prior is
\begin{equation}
\alpha_\ell(\Delta)
=
1-2^{-\Delta/h_\ell}.
\label{eq:hierarchical_memory_prior}
\end{equation}
Shorter-timescale levels therefore place a larger prior weight on a new
segment, whereas longer-timescale levels change more conservatively.

The prior is combined with a content-dependent correction computed from the
existing slot and its attention candidate:
\begin{equation}
\begin{aligned}
g_{\ell,k}
&=
\operatorname{sigmoid}
\left(
\operatorname{logit}\!\left(\alpha_\ell(\Delta)\right)
+
G\!\left([M_{\ell,k};\widetilde M_{\ell,k}]\right)
\right),\\
M'_{\ell,k}
&=
\operatorname{LN}
\left(
M_{\ell,k}
+
g_{\ell,k}\odot
(\widetilde M_{\ell,k}-M_{\ell,k})
\right),
\end{aligned}
\label{eq:hierarchical_memory_update}
\end{equation}
where $g_{\ell,k}\in(0,1)^d$ provides a feature-wise gated interpolation.
Consequently, the temporal prior determines the nominal update rate of each
level, while the learned correction allows the content of a particular
overflow segment to modulate that update.

The long-term branch is itself temporally bounded. Only a fixed number of the
most recent overflow segments, determined by the largest configured memory
timescale, contribute to the persistent state. Writing this bounded ordered set
as $\mathcal Q_c$, the effective memory can be viewed as applying
Eq.~\ref{eq:hierarchical_memory_update} sequentially over
$\mathcal Q_c$ in temporal order. Hence both the number of explicit recent
tokens and the number of long-term memory slots remain independent of episode
length.

At readout, the long-term slots are flattened and concatenated with the
temporally encoded recent segments to form the persistent history supplied to
the next cycle:
\begin{equation}
H_c=[M_c;R_c].
\label{eq:app_persistent_history}
\end{equation}
This construction preserves explicit access to the most recent realized
interaction while progressively summarizing older executed context at multiple
temporal scales. Predicted but unexecuted observations never enter either
branch of the persistent history.

\subsection{Block-Causal Multimodal Computation}
\label{app:method_visibility}

\begin{figure*}[t]
    \centering
    \includegraphics[width=\textwidth]{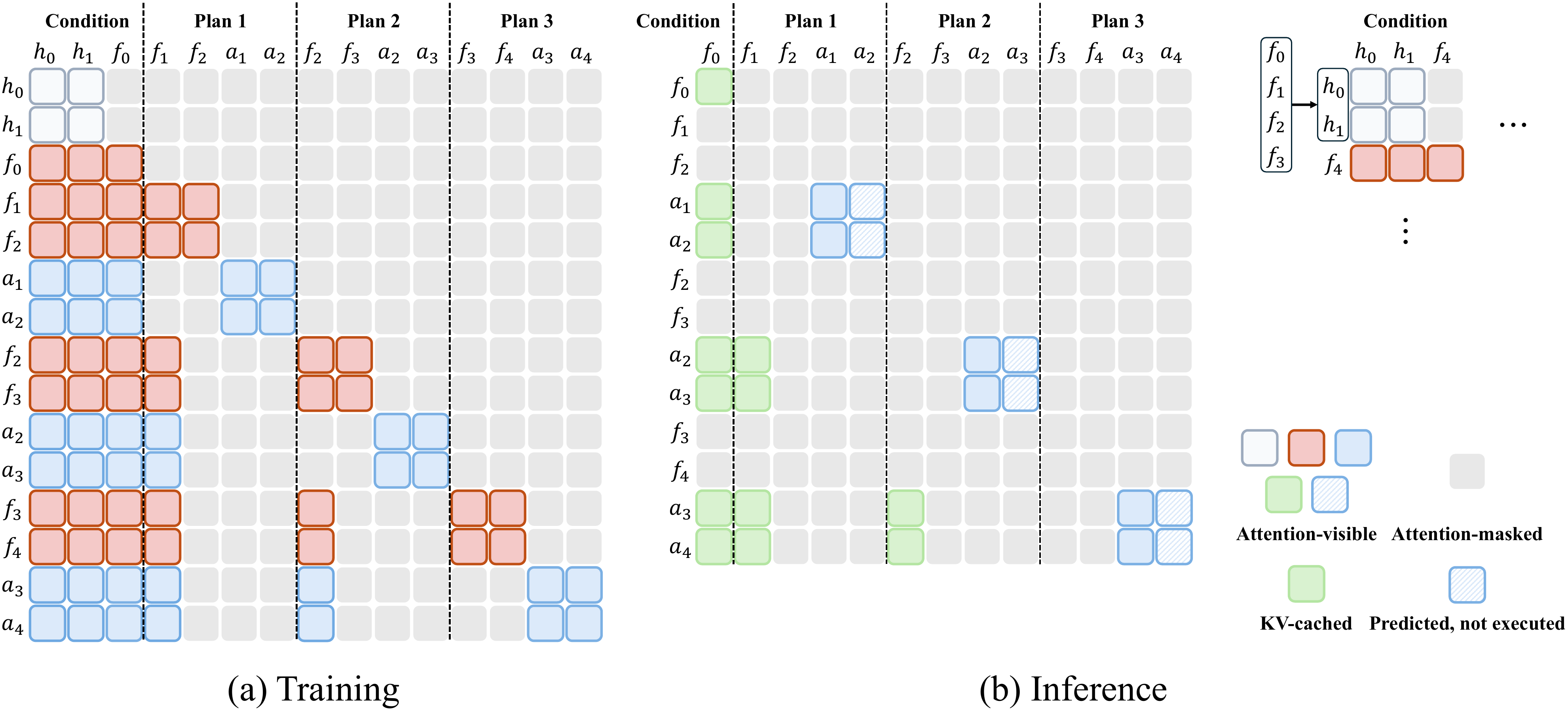}
    \caption{
    \textbf{Block-causal visibility for the default modality-decoupled
    world--action formulation.}
    The training view (left) and autoregressive rollout view (right) show
    three overlapping plans. Visual and action predictions receive the same
    clean causal prefix, while simultaneously corrupted future visual and
    action tokens are not used as cross-modal conditions. Completed blocks
    are teacher-forced during training; at rollout, only observations reached
    through committed execution enter the clean prefix. Planned but
    unexecuted suffixes are discarded before the next decision. Colored cells
    indicate visible entries, gray cells are masked, green cells denote clean
    retained context, and hatched cells denote predicted but unexecuted
    variables.}
    \label{fig:mask_decoupled}
\end{figure*}

\begin{figure*}[t]
    \centering
    \includegraphics[width=\textwidth]{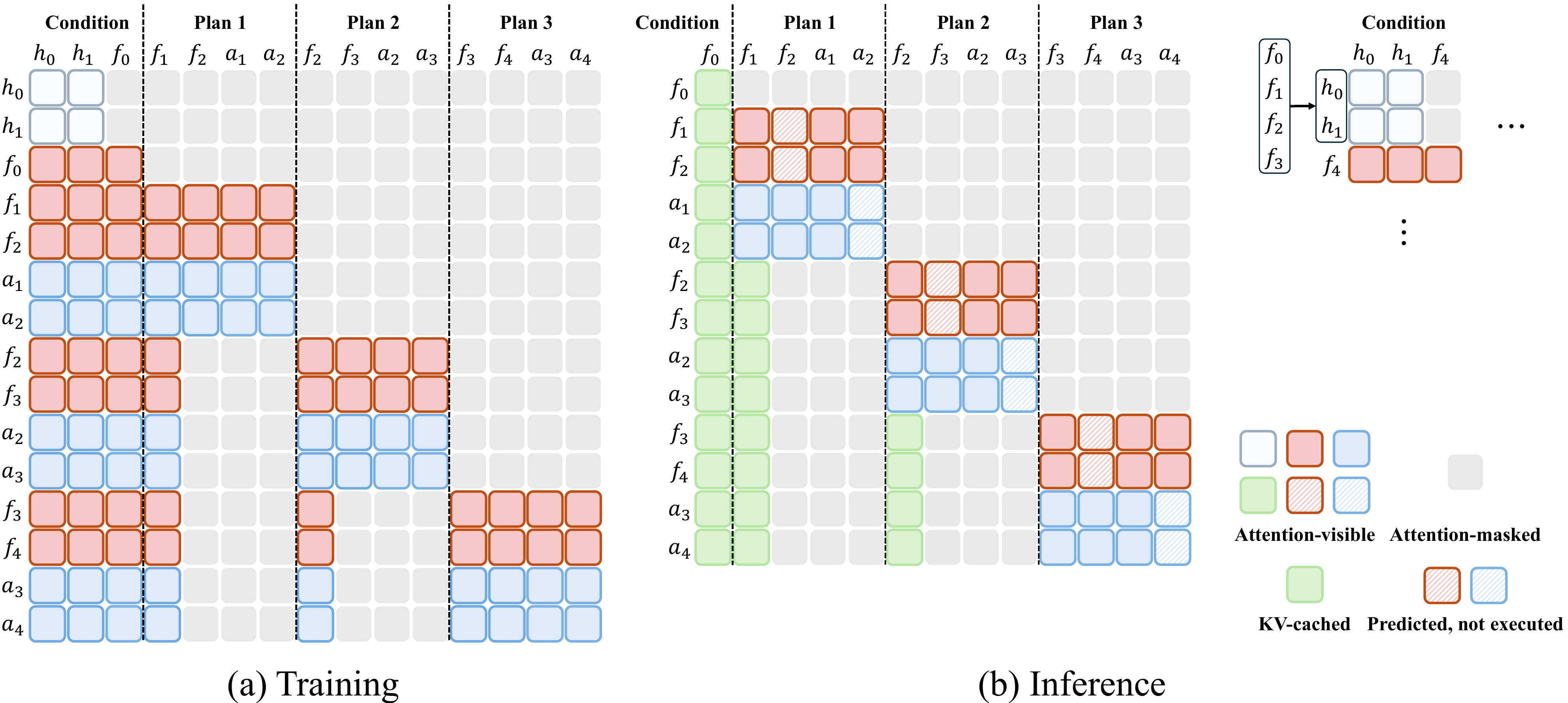}
    \caption{
    \textbf{Block-causal visibility with joint current-block video--action
    coupling.}
    The joint alternative allows aligned noisy visual and action variables
    to exchange information inside the current planning block while retaining
    the same clean causal prefix and the same exclusion of later blocks.
    Training and rollout otherwise follow the same teacher-forcing and
    execution-grounded context rules as in \cref{fig:mask_decoupled}.}
    \label{fig:mask_coupled}
\end{figure*}

The main text introduces the modality-decoupled block-causal mask used by the
reported GameWAM model. Here we make its transformer-level realization
explicit and contrast it with a joint within-block video--action alternative.
For target block $j$, let $\mathcal P_{c,j}$ denote the causally available
clean visual prefix, and let $\mathcal V_j$ and $\mathcal A_j$ denote the
corresponding noisy video and action variables. Instruction, proprioceptive
context, and other non-prefix conditions remain available through their
respective conditioning paths and are omitted below for clarity.

At the transformer level, the Video and Action DiTs retain modality-specific
hidden states, prediction parameters, and flow-time modulation. Their
interaction is instead determined by the key/value states made visible to each
query stream. For modality $m\in\{v,a\}$ at layer $\ell$, a generic update can
be written as
\begin{equation}
\begin{aligned}
\widetilde h^{m,\ell}
&= h^{m,\ell}
+ G^{m,\ell}_{\mathrm{attn}}
\!\left(
\operatorname{Attn}
\left(
Q^{m,\ell},
\mathbf K^{m,\ell}_{c,j},
\mathbf V^{m,\ell}_{c,j}
\right);
\sigma_m
\right),\\
h^{m,\ell+1}
&= \widetilde h^{m,\ell}
+ G^{m,\ell}_{\mathrm{ffn}}
\!\left(
\widetilde h^{m,\ell};
\sigma_m
\right),
\end{aligned}
\label{eq:app_transformer_update}
\end{equation}
where the modality-specific gates absorb the corresponding flow-time
modulation, and
$(\mathbf K^{m,\ell}_{c,j},\mathbf V^{m,\ell}_{c,j})$ denotes the
layer-$\ell$ attention context assembled according to the block-causal
visibility rule.

For the default modality-decoupled mask,
\begin{equation}
\operatorname{Vis}(\mathcal A_j)
=
\mathcal P_{c,j}\cup\mathcal A_j,
\qquad
\operatorname{Vis}(\mathcal V_j)
=
\mathcal P_{c,j}\cup\mathcal V_j.
\label{eq:app_decoupled_mask}
\end{equation}
Importantly, this decoupling applies to the simultaneously corrupted
\emph{future} variables; it does not separate action generation from the
learned representation of the realized visual world.

Concretely, at each transformer layer the complete clean visual prefix
$\mathcal P_{c,j}$ is processed by the Video DiT to produce layer-wise
clean-prefix key/value states
$(\mathbf K^{p,\ell}_{c,j},\mathbf V^{p,\ell}_{c,j})$. The same states are
reused by both prediction branches. Let
$(\mathbf K^{v,\ell}_j,\mathbf V^{v,\ell}_j)$ and
$(\mathbf K^{a,\ell}_j,\mathbf V^{a,\ell}_j)$ denote the key/value states of
the current noisy video and action variables, respectively. Omitting the
additional conditioning paths above, the default attention contexts have the
schematic form
\begin{equation}
\begin{aligned}
\mathbf K^{v,\ell}_{c,j}
&=
\left[
\mathbf K^{p,\ell}_{c,j};
\mathbf K^{v,\ell}_j
\right],
&
\mathbf V^{v,\ell}_{c,j}
&=
\left[
\mathbf V^{p,\ell}_{c,j};
\mathbf V^{v,\ell}_j
\right],
\\
\mathbf K^{a,\ell}_{c,j}
&=
\left[
\mathbf K^{p,\ell}_{c,j};
\mathbf K^{a,\ell}_j
\right],
&
\mathbf V^{a,\ell}_{c,j}
&=
\left[
\mathbf V^{p,\ell}_{c,j};
\mathbf V^{a,\ell}_j
\right].
\end{aligned}
\label{eq:app_decoupled_attention_context}
\end{equation}
Thus, video queries attend to the realized clean prefix and the current noisy
video states, whereas action queries attend to the same realized clean prefix
and the current noisy action states. The action branch does not read
current-block noisy-video K/V states, and the video branch does not read
current-block noisy-action K/V states. The two branches therefore remain
coupled through a shared learned representation of the realized clean prefix,
rather than through direct exchange between simultaneously denoised future
variables.

This distinction is also important for understanding the role of video
supervision. The clean-context K/V states are produced by the Video DiT and
remain in the training computation graph when they are consumed by both
branches. Consequently, gradients from both the video and action objectives
propagate through the Video DiT path that constructs this shared context.
Video prediction can therefore shape the realized-world representation used by
action generation even though the action prediction does not condition on a
denoised future video. In this sense, the default formulation couples world
and action learning through the shared representation of realized dynamics,
while deliberately decoupling their current noisy future variables.

\cref{fig:mask_decoupled} visualizes this computation under
teacher-forced training and autoregressive rollout. During training, completed
blocks provide clean causal context, but future targets that are not yet
causally available remain excluded from the corresponding clean prefix. During
online interaction, only observations reached after committed environment
actions are promoted into the realized context; predicted but unexecuted
suffixes never become clean conditioning states.

The same dependency structure is retained during action-only online inference.
The instantiated clean prefix $\mathcal P_{c,j}$ is processed by the Video DiT
to construct layer-wise clean-prefix K/V states, which are cached and consumed
by Action DiT denoising. As new observations are obtained through execution,
only the newly realized within-cycle context extends this cache. What is
omitted during action-only inference is the iterative denoising of
\emph{future} video variables, not the Video DiT encoding of the realized
clean prefix. The default mask therefore preserves the learned visual-context
pathway used for action generation while avoiding the additional online cost
of future-video denoising.

For the joint within-block alternative,
\begin{equation}
\operatorname{Vis}(\mathcal A_j)
=
\operatorname{Vis}(\mathcal V_j)
=
\mathcal P_{c,j}\cup\mathcal V_j\cup\mathcal A_j.
\label{eq:app_joint_mask}
\end{equation}
The corresponding attention contexts contain both current noisy modalities:
\begin{equation}
\begin{aligned}
\mathbf K^{v,\ell}_{c,j}
=
\mathbf K^{a,\ell}_{c,j}
&=
\left[
\mathbf K^{p,\ell}_{c,j};
\mathbf K^{v,\ell}_j;
\mathbf K^{a,\ell}_j
\right],
\\
\mathbf V^{v,\ell}_{c,j}
=
\mathbf V^{a,\ell}_{c,j}
&=
\left[
\mathbf V^{p,\ell}_{c,j};
\mathbf V^{v,\ell}_j;
\mathbf V^{a,\ell}_j
\right].
\end{aligned}
\label{eq:app_joint_attention_context}
\end{equation}

The joint formulation therefore introduces direct bidirectional interaction
between the current noisy future-video and action variables during denoising.
The temporal causal boundary is otherwise unchanged: both formulations use
the same realized clean prefix and exclude information from later blocks.

This additional within-block coupling also changes online computation. Because
the action queries depend on the simultaneously denoised future-video states,
joint action generation requires the future visual stream to be denoised
alongside the action stream. By contrast, the default modality-decoupled model
retains the shared realized-context pathway while allowing future-video
denoising to be omitted during action-only deployment. We compare the two
formulations under matched training and evaluation settings in
\cref{app:mask_comparison}.

\subsection{Interaction-Conditioned Native Action Generation}
\label{app:interaction_conditioned_action}

Minecraft gameplay and GUI interaction use the same physical keyboard--mouse
interface but induce substantially different conditional control
distributions. GameWAM therefore retains a single native action
representation while specializing action prediction by interaction mode. The
specialization is applied independently at each native action timestep rather
than once to an entire prediction horizon, allowing a generated trajectory to
transition between gameplay and GUI interaction without switching models or
invoking a separate high-level controller.

The Action DiT produces gameplay- and GUI-specific flow predictions together
with a timestep-wise routing prediction. During training, the observed
interaction-mode label selects which action branch is supervised at each
timestep and simultaneously provides the target for the routing loss. During
rollout, the predicted route instead selects the branch used to generate the
complete native action vector at that timestep. Routing therefore chooses
between interaction-specific conditional predictions while preserving a shared
action dimensionality and a single low-level control interface.

This distinction is particularly important for continuous control. The same
continuous coordinates represent camera motion during gameplay and cursor
motion during GUI interaction, but their numerical distributions differ
substantially. We therefore normalize continuous action coordinates with
interaction-specific statistics before flow modeling and invert the
corresponding normalization after generation. Binary keyboard and mouse
coordinates retain their shared normalization, while validity masks exclude
coordinates unavailable for a particular environment or interaction state.
Environment-specific thresholding, hotbar constraints, and coordinate
definitions are given in \cref{app:action_space}.

The router changes the conditional distribution used to predict an action; it
does not decompose the controller into independent gameplay and GUI policies.
Both branches operate on the same Action DiT representation and predict the
same native action vector, and the route can change from one action timestep
to the next. Consequently, heterogeneous interaction modes are handled within
a single world--action trajectory while retaining mode-appropriate continuous
control statistics.

\subsection{Supervision Semantics Beyond the Compact Objective}
\label{app:supervision_semantics}

The losses in \cref{sec:training_objective} already define the optimized
quantities. Three details clarify how those terms are applied. First, valid
continuous and binary action coordinates are reduced separately before being
combined. This normalizes heterogeneous action coordinates but does not imply
independent action policies; both groups belong to the same generated native
action trajectory. Validity masks exclude padding and coordinates unavailable
in a particular environment or interaction state.

Second, teacher forcing applies clean context only to blocks that are causally
complete at the prediction boundary. The current block's future visual and
action variables remain generation targets and are corrupted independently,
while completed visual blocks provide clean conditioning. This preserves the
realized-versus-uncertain partition used by online rollout.

Finally, the predictive-history objective in
Eq.~\plaineqref{eq:history_loss} operates only on the bounded cross-cycle
history. Its pooled representation predicts the clean visual feature at the
current observation boundary, which is treated as a fixed target. This
auxiliary constraint discourages information loss during history compression
and encourages $H_c$ to retain visual information predictive across cycle
boundaries, without changing the main world--action flow objective. Numerical
weights are given in \cref{app:experimental_protocol}.

\section{Data and Training Construction}
\label{app:data_details}

\subsection{Trajectory Alignment}

All data sources are converted to a common observation--state--action
timeline. The state paired with an action is the state observed immediately
before that action is executed. Binary controls, continuous camera/cursor
values, padding indicators, and validity masks for unavailable state or action
fields are retained before normalization. The same pre-action convention
determines whether a Minecraft step belongs to gameplay or GUI interaction.

For VPT, video observations are synchronized with the corresponding per-frame
mouse, keyboard, and Minecraft state records before conversion to the shared
trajectory representation. A null-action filtering stage removes redundant
null controls together with their paired frames while preserving alignment
between retained observations, state, actions, and event metadata. The
resulting episodes store aligned action and state records, episode-aligned
visual observations, and the metadata required for training.

\subsection{Native Action Interfaces}
\label{app:action_space}

The main text defines GameWAM over hybrid environment actions.
\cref{tab:app_mc_action_space,tab:app_vizdoom_action_space}
summarize the action coordinates used in our experiments.

\begin{table}[t]
\centering
\small
\caption{\textbf{Minecraft action interface.} The first two coordinates are
continuous camera deltas; the remaining coordinates are binary keyboard or
mouse controls.}
\label{tab:app_mc_action_space}
\begin{tabular}{rll@{\hspace{1.7em}}rll}
\toprule
Index & Control & Type & Index & Control & Type \\
\midrule
0  & Camera pitch  & continuous & 11 & Drop item       & binary \\
1  & Camera yaw    & continuous & 12 & Open inventory  & binary \\
2  & Move forward  & binary     & 13 & Hotbar slot 1   & binary \\
3  & Move backward & binary     & 14 & Hotbar slot 2   & binary \\
4  & Move left     & binary     & 15 & Hotbar slot 3   & binary \\
5  & Move right    & binary     & 16 & Hotbar slot 4   & binary \\
6  & Jump          & binary     & 17 & Hotbar slot 5   & binary \\
7  & Sneak         & binary     & 18 & Hotbar slot 6   & binary \\
8  & Sprint        & binary     & 19 & Hotbar slot 7   & binary \\
9  & Attack        & binary     & 20 & Hotbar slot 8   & binary \\
10 & Use/interact  & binary     & 21 & Hotbar slot 9   & binary \\
\bottomrule
\end{tabular}
\end{table}

\begin{table}[t]
\centering
\small
\caption{\textbf{ViZDoom action interface used in our experiments.} A unified
9-D representation covers both map families. Battle maps use continuous
horizontal turning, whereas the two defense maps use binary left/right
turning.}
\label{tab:app_vizdoom_action_space}
\begin{tabular}{rlll}
\toprule
Index & Control & Type & Used in \\
\midrule
0 & Fire                  & binary     & all maps \\
1 & Move forward          & binary     & Battle maps \\
2 & Move backward         & binary     & Battle maps \\
3 & Strafe left           & binary     & Battle maps \\
4 & Strafe right          & binary     & Battle maps \\
5 & Speed modifier        & binary     & Battle maps \\
6 & Turn left             & binary     & defense maps \\
7 & Turn right            & binary     & defense maps \\
8 & Horizontal turn delta & continuous & Battle maps \\
\bottomrule
\end{tabular}
\end{table}

The camera coordinates are normalized continuously. Binary controls are
thresholded after generation, and the hotbar coordinates are constrained to
select at most one slot. Gameplay and GUI interaction share the same action
layout but use separate normalization statistics and mode-specific decoding.
The Minecraft state input contains view orientation, cursor location, selected
hotbar slot, and GUI state; unavailable fields are explicitly masked.

For Battle maps, the continuous turn coordinate is normalized to $[-1,1]$,
corresponding to a horizontal rotation of up to $10^\circ$ per interaction
step. No separate continuous pitch control is used in the reported ViZDoom
experiments. The proprioceptive input contains planar position, heading, and
health, with unavailable fields treated as missing rather than observed zeros.

\begin{figure*}[t!]
    \centering
    \includegraphics[width=\textwidth]
    {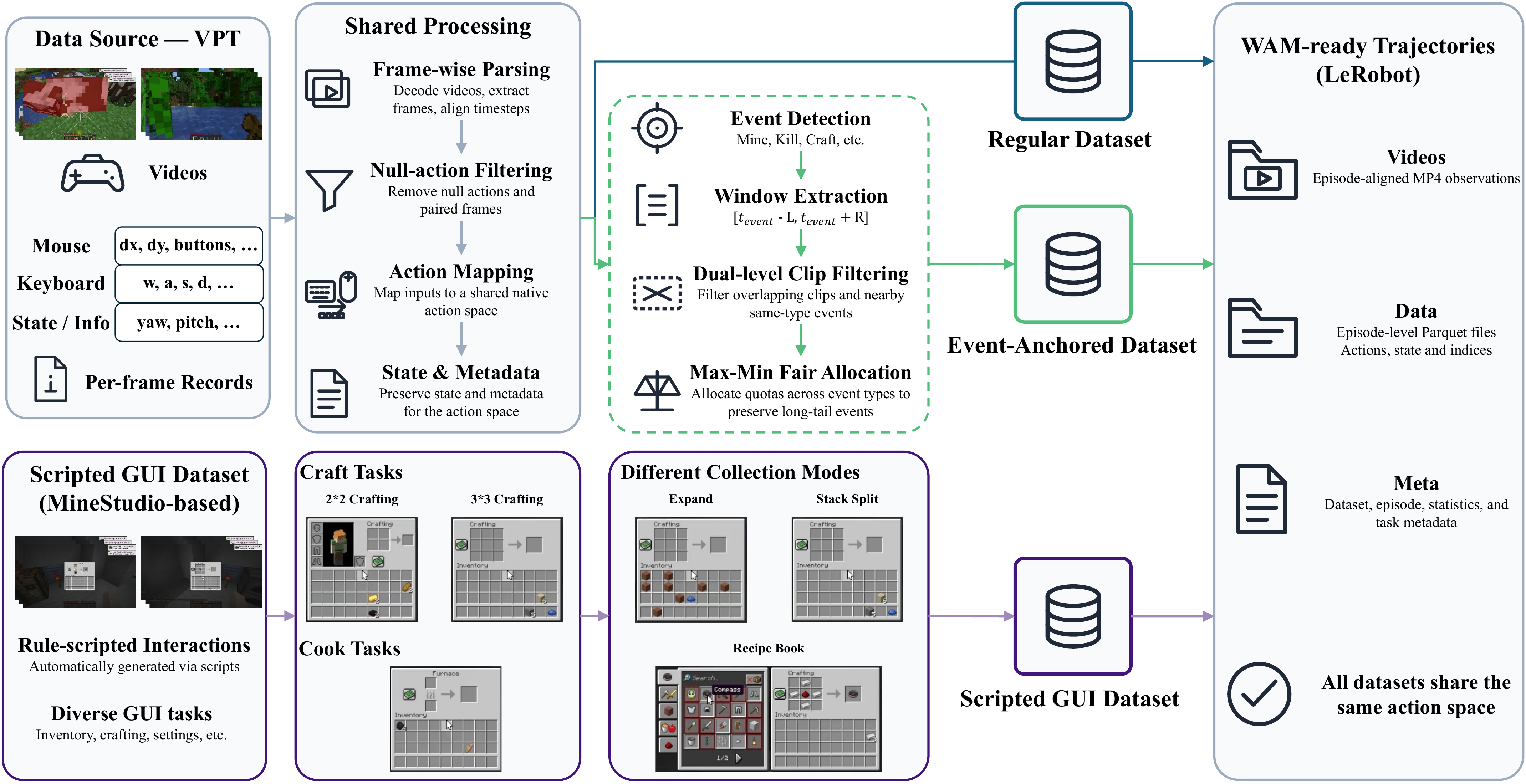}
    \caption{
    \textbf{Construction of WAM-ready game trajectories.}
    VPT recordings are standardized into regular and Event-Anchored datasets;
    the event branch detects interaction events, extracts anchored windows,
    filters temporally redundant candidates, and applies max--min fair
    allocation across event types. MineStudio-based scripted GUI trajectories
    broaden interface-interaction coverage. All streams share the same native
    action space and trajectory representation for joint world--action
    training.
    }
    \label{fig:app_data_pipeline}
\end{figure*}

\subsection{Training Data Sources}

\paragraph{Minecraft.}

Minecraft training uses three complementary streams. Regular VPT
trajectories~\citep{baker2022vpt} provide broad naturalistic interaction
coverage and long temporal context. From the same source recordings, we
construct an Event-Anchored VPT dataset around identifiable MineStudio-style
interaction events~\citep{MineStudio}, providing instruction-conditioned
supervision around behaviorally meaningful transitions. MineStudio-based
scripted GUI trajectories form the third stream and broaden
interface-interaction coverage. All three streams follow the same
observation--state--action alignment and native action representation.

The two VPT streams differ at the dataset level. Regular VPT preserves the
standardized long trajectories, whereas the Event-Anchored stream applies the
offline event-selection procedure described below to construct anchored
sub-trajectories. Event-anchored clip sampling is subsequently applied during
training and is therefore distinct from construction of the Event-Anchored
Dataset.

\paragraph{ViZDoom.}

We collect expert trajectories using Asynchronous Proximal Policy Optimization
(APPO)~\citep{petrenko2020samplefactory} on the two Battle maps and the Defend
the Center and Defend the Line maps. The action actually executed at each
recorded interaction step is preserved and expressed in the unified 9-D
representation of \cref{tab:app_vizdoom_action_space}. Continuous horizontal
turning is retained on the Battle maps rather than discretized. ViZDoom does
not require a separate gameplay/GUI interaction mode.

\subsection{Event-Anchored VPT Dataset Construction}
\label{app:event_dataset}

The Event-Anchored VPT dataset is constructed offline from the same source
recordings as Regular VPT. Its construction proceeds through event detection,
anchored window extraction, temporal redundancy filtering, and allocation
across event types. The concrete construction parameters used for the reported
dataset are provided with the sampling configuration in
\cref{app:training_sampling_config}.

\paragraph{Event detection.}
We identify MineStudio-style interaction events from changes in the recorded
Minecraft state statistics. The retained event set covers identifiable
interactions such as mining, entity interaction, crafting, and related state
transitions. Each detected event is associated with its temporal location and
corresponding instruction metadata, providing an anchor aligned with the
original visual and native-action trajectory.

\paragraph{Anchored window extraction.}
For an event occurring at temporal index $t_i$, we extract a surrounding
trajectory window

\begin{equation}
W_i = X[t_i-L:t_i+R],
\label{eq:event_window}
\end{equation}

where $X$ denotes the aligned observation--state--action trajectory and $L$
and $R$ specify the retained context around the event. The event position is
preserved within the extracted sub-trajectory as anchor metadata used by the
subsequent filtering and training-sampling stages.

\paragraph{Dual-level clip filtering.}
Event detection can yield temporally redundant candidates, either because
different detected events produce strongly overlapping windows or because
multiple occurrences of the same event type appear in close succession. We
therefore apply two temporal filtering criteria: one limits overlap among
candidate clips, while the other suppresses nearby repetitions of the same
event type. This reduces redundant supervision before event-type allocation
while retaining distinct interactions and their surrounding context.

\paragraph{Max--min fair allocation.}
The filtered candidate pool remains imbalanced because event types occur with
substantially different frequencies in natural trajectories. We therefore
allocate the retained dataset budget across event types using max--min
fairness. Event types with fewer available candidates saturate at their
available count, while the remaining capacity is redistributed among event
types with larger candidate pools. This limits domination by highly frequent
events while preserving long-tail interactions whenever sufficient candidates
are available.

The retained windows, together with their event anchors, instruction metadata,
aligned observations, state, and native actions, constitute the
Event-Anchored VPT dataset.

\subsection{Scripted GUI Trajectory Construction}
\label{app:gui_data}

The scripted GUI stream is generated in MineStudio through rule-scripted
native keyboard--mouse interaction. The collection covers diverse interface
procedures and uses multiple collection modes to vary the interaction
trajectories. The scripts are used only to generate demonstrations: the
resulting training data contains rendered observations, proprioceptive state,
and native actions in the same representation used by the VPT streams.

The collected episodes undergo the same frame-wise parsing, state preservation,
action mapping, and trajectory alignment as the other Minecraft data sources.
Consequently, scripted GUI data broadens interface-interaction coverage without
introducing a separate action representation for training.

\subsection{Event-Anchored Training Clip Sampling}
\label{app:event_anchor_sampling}

Event-Anchored Dataset construction determines which anchored
sub-trajectories are retained, whereas event-anchored clip sampling separately
determines the temporal density of training windows drawn from them.

Long interaction trajectories contain highly uneven supervision density.
Transitions near an annotated event often contain the visual changes and
actions most directly related to the corresponding instruction, whereas more
distant portions may consist largely of traversal, incidental camera motion,
or other weakly related behavior. Uniform window sampling can therefore devote
substantial training capacity to comparatively less informative regions.

For the Event-Anchored VPT stream, we increase the density of candidate
training windows whose observation span overlaps the neighborhood of the
stored event anchor, while sampling the remaining regions more sparsely. This
increases exposure to event-relevant state transitions while retaining
surrounding interaction context. Regular VPT and scripted GUI trajectories use
fixed sampling intervals rather than anchor-dependent sampling. The mixture
weights, anchor neighborhood, and clip-sampling intervals used in the reported
training run are given in \cref{app:training_sampling_config}.

\section{Implementation and Evaluation Protocol}
\label{app:experimental_protocol}

\subsection{Model Architecture and Initialization}

The visual predictor is initialized from Wan2.2-TI2V-5B. The Action DiT
follows the same 30-layer, 24-head transformer layout but uses a reduced
hidden width of 1024, compared with 3072 for the Video DiT. To retain the
pretrained video-transformer prior despite this width reduction, we adapt the
Wan2.2 Video DiT weights to initialize the Action DiT backbone, while
action-specific input/output projections, gameplay/GUI prediction heads, and
the interaction-mode router are initialized separately and learned during
GameWAM training. Diffusion-time modulation is applied independently to the
two streams.

The video VAE is kept frozen, and text-conditioning features are produced by a
fixed text encoder. The Video DiT, Action DiT (including the interaction-specific
action prediction heads and router), hierarchical history module, and
proprioceptive encoder are optimized during GameWAM training.

\subsection{Observation, Planning, and Temporal-Context Configuration}

Both environments use $224\times224$ RGB observations. Visual observations are
sampled once every two native environment actions. The prediction horizon is
$P=16$ native actions and the execution horizon is $E=8$, giving an eight-action
overlap between neighboring predictions. Three committed execution blocks form
one 24-action cycle for the hierarchical temporal context.

For each completed cycle, the cross-cycle representation preserves four latent
time positions and projects each position to four spatial tokens, giving 16
tokens per cycle-level segment. The two most recent completed segments are kept
explicitly. Older context is summarized at two temporal scales with four memory
tokens per scale; the corresponding scale half-lives are two and four completed
cycles. The resulting cross-cycle history contains at most 40 tokens. During
training, the available cross-cycle history is shortened with probability
$0.1$ by retaining a randomly selected valid suffix, including the possibility
of no cross-cycle history.

The environment-specific action interfaces are given in
\cref{tab:app_mc_action_space,tab:app_vizdoom_action_space}.
Minecraft uses timestep-wise interaction conditioning for gameplay and GUI
behavior, whereas the evaluated ViZDoom tasks use a single interaction regime.

\subsection{Training Data Mixture and Clip-Sampling Configuration}
\label{app:training_sampling_config}

\noindent\textbf{Event-Anchored VPT construction.}
The reported Event-Anchored VPT dataset uses 96-frame sub-trajectories. For an
event at frame $t$, the extraction window ends eight frames after the event,
giving the interval $[t-87,t+8]$ and retaining both substantial pre-event
context and a short post-event continuation. Null-action filtering is applied
before event-window construction while preserving detected event anchors.
During dual-level filtering, retained clip starts are separated by at least
64 frames, and anchors of the same event type are separated by more than
96 frames. The filtered candidates from the VPT partitions are pooled and
subject to a global max--min fair allocation with a total budget of 200,000
Event-Anchored sub-trajectories. Candidate selection is shuffled with random
seed 42. VPT data are processed at their native 20 FPS.

\noindent\textbf{Training mixture and temporal sampling.}
The reported Minecraft training run uses sample-mixture weights of 80\% for
Event-Anchored VPT, 5\% for Regular VPT, and 15\% for scripted GUI
trajectories. These weights specify the training-sample distribution rather
than the raw corpus composition.

The three streams use different temporal intervals for constructing training
clips. For Event-Anchored VPT, candidate windows whose observation span
intersects a radius-$8$ neighborhood of the stored event anchor are sampled
with stride 2, while the remaining regions use stride 16. Regular VPT uses
stride 32 throughout the trajectory, whereas scripted GUI trajectories use
stride 8. Thus, only the Event-Anchored stream varies its training-window
density according to temporal proximity to an interaction event.

\subsection{Optimization and Objective Weighting}

GameWAM is trained for two epochs using fused AdamW with
$\beta_1=0.9$, $\beta_2=0.95$, weight decay $0.01$, and gradient clipping at a
maximum norm of $1.0$. The learning rate is linearly warmed up over the first
5\% of optimization, corresponding to 1,095 steps, to a peak value of
$4\times10^{-5}$, and then follows cosine decay to a minimum of
$4\times10^{-7}$. Each epoch contains 10,950 optimizer steps, for 21,900 steps
in total.

Training uses BF16 precision and DeepSpeed ZeRO-2 on eight NVIDIA H200 GPUs.
The per-GPU batch size is 44 with gradient accumulation of one, giving an
effective global batch size of 352. The full two-epoch training run takes
approximately 22 hours.

For the objective in Eq.~\plaineqref{eq:total_objective}, the reported model
uses $\lambda_v=1.0$, $\lambda_a=1.0$, $\lambda_m=0.05$, and
$\lambda_h=0.5$. Within the grouped action objective in
Eq.~\plaineqref{eq:grouped_action_flow_loss}, continuous and discrete action
losses are weighted equally with
$\lambda_{\mathrm{cont}}=\lambda_{\mathrm{disc}}=1.0$.

\subsection{Flow-Training and Sampling Configuration}

The visual and action flow times are sampled independently. For
$u\sim\mathcal U[0,1]$, the reported training runs use the shifted time mapping
\begin{equation}
    \sigma=\frac{5u}{1+4u}.
    \label{eq:app_shifted_sigma}
\end{equation}
The corresponding flow residuals are weighted by
\begin{equation}
    w(\sigma)
    =\frac{
       \exp[-2(\sigma-\tfrac12)^2]-\exp(-\tfrac12)
    }{Z},
    \label{eq:app_flow_weight}
\end{equation}
where $Z$ normalizes the mean weight under the sampling schedule. At inference,
sampling starts from Gaussian sources and follows the learned vector field from
$\sigma=1$ to $\sigma=0$ with first-order integration. Standard closed-loop
evaluation uses ten denoising steps.

\subsection{Closed-Loop Evaluation Protocols}
\label{app:closed_loop_protocols}

All reported task-performance evaluations use the corresponding final
checkpoint after the second training epoch, without validation- or MCU-based
checkpoint selection. Unless otherwise stated, the action source is
independently resampled at each replanning step rather than reused across an
episode. This mitigates coherent episode-level accumulation from repeated
reuse of a single source; the corresponding LASI failure mode and
source-resampling analysis are detailed in \cref{app:lasi}.

\paragraph{Minecraft MCU.}

We use the MCU protocol reported in the main paper, including the Mini subset
of 30 tasks (10 mining, 10 crafting, and 10 combat tasks). Success is
determined by the environment's task-completion signals. Interaction steps
count native environment actions actually executed during closed-loop rollout,
excluding internal denoising evaluations. The reported step metric is averaged
over successful episodes within each task category and is undefined when no
successful episode is observed. The reported task-wise standard deviation is
computed across per-task success rates. Except for the reproduced Fast-WAM comparison, the remaining systems in
\cref{tab:mcu} differ substantially in training data, task-specific
supervision, model class, action representation, and available inputs.
Their results should therefore be interpreted primarily as a system-level
benchmark comparison rather than a controlled comparison of architecture
or data efficiency.

\paragraph{ViZDoom.}

We use the four-map protocol reported in the main paper and measure average
episode reward. The controller uses the action interface in
\cref{tab:app_vizdoom_action_space} and the same $P=16$, $E=8$
overlapping-plan geometry as Minecraft.

\paragraph{Online execution frequency.}

For the inference-efficiency comparison in
\cref{app:mask_comparison} and \cref{tab:mask_comparison}, online execution frequency is evaluated
separately from task-performance evaluation on a single NVIDIA H200 GPU under
the standard inference configuration. For each model variant, we report the
mean frequency over 10 closed-loop episodes. Timing includes all model-side
computation performed during online replanning and excludes environment
simulation and execution time.

\section{Additional Experimental Analysis}
\label{app:additional_analysis}

\subsection{Training Scale and Token Consumption}
\label{app:token_consumption}

\cref{tab:training_tokens} places the training scale of GameWAM in
context with OpenHA~\citep{wang2026openha} and
Game-TARS~\citep{wang2025gametars}. We retain the token categories reported
by each method rather than forcing heterogeneous training pipelines into a
shared modality definition. GameWAM uses a single training stage, whereas
OpenHA and the reported Game-TARS training recipe use two stages.

\begin{table}[t]
\centering
\small
\setlength{\tabcolsep}{5.0pt}
\renewcommand{\arraystretch}{1.12}
\caption{
\textbf{Model scale and reported training-token consumption.}
Token categories follow the accounting reported by each work. For two-stage
methods, modality-level counts are shown as Stage~1 / Stage~2. A dash indicates
that the corresponding count is not separately reported or is not applicable.
}
\label{tab:training_tokens}
\resizebox{\linewidth}{!}{
\begin{tabular}{lccc}
\toprule
&
\textbf{GameWAM} &
\textbf{OpenHA} &
\textbf{Game-TARS} \\
\midrule

\multicolumn{4}{l}{\textit{Model initialization and scale}} \\

Backbone(s) &
\begin{tabular}[c]{@{}c@{}}
Video: Wan2.2-TI2V-5B \\
Action: Wan2.2-initialized Action DiT
\end{tabular}
&
Qwen2-VL-7B
&
Qwen2.5-VL-7B-Instruct
\\

Backbone scale / params. &
\begin{tabular}[c]{@{}c@{}}
Video: 5B \\
Action DiT: 1B
\end{tabular}
&
7B
&
7B
\\

\midrule
\multicolumn{4}{l}{\textit{Training-token exposure}} \\

Stage 1 &
\multirow{2}{*}{\textit{Single stage}} &
3.40B &
526B \\

Stage 2 &
&
0.22B &
40B \\

\midrule

Video tokens &
2.27B &
-- &
-- \\

Image tokens &
-- &
-- &
208B / 10B \\

Action tokens &
0.37B &
-- &
-- \\

Text tokens &
0.15B &
-- &
326B / 29B \\

\midrule

\textbf{Total consumed tokens} &
\textbf{2.79B} &
\textbf{3.62B} &
\textbf{566B} \\

\bottomrule
\end{tabular}
}
\end{table}

For GameWAM, the reported counts follow the exact two-epoch sampling schedule
rather than an estimate based on average sequence length. Video and action
tokens count target-token presentations participating in the training
objective, including overlapping autoregressive planning targets. Text tokens
count non-padding tokenizer tokens under the same sampling schedule;
zero-padded text slots are excluded. Across the two epochs, GameWAM consumes
2.27B video-target tokens, 0.37B action-target tokens, and 0.15B effective
text-conditioning tokens, for 2.79B total consumed tokens.

The exact schedule presents 7,708,800 training samples over the two epochs.
The 150,723,927 non-padding text tokens therefore correspond to an average of
19.55 effective text tokens per presented sample. The implementation uses a
fixed 128-slot text context, corresponding to 986,726,400 text slots over the
same schedule, but most of the additional positions are zero padding. We
therefore use non-padding tokenizer tokens for the reported text-token
consumption rather than counting padded computational slots.

For OpenHA, Stage~1 and Stage~2 correspond to its mixed-action pre-training
and CoA fine-tuning stages, respectively. The reported consumed-token counts
are 3.40B and 0.22B, giving 3.62B in total. A modality decomposition directly
comparable to the video, action, and text accounting used for GameWAM is not
separately reported, so the corresponding entries are left unspecified.

For Game-TARS, we reproduce the stage-wise quantities in its reported training
recipe without redefining their token accounting. Pre-training reports 526B
total tokens, including 208B image tokens and 326B text tokens, while
post-training reports 40B total tokens, with 10B image tokens and 29B text
tokens. We retain these values exactly as reported rather than reconstructing
the totals from the modality columns. Our experimental comparison uses the
Game-TARS-Dense model, whose reported initialization is
Qwen2.5-VL-7B-Instruct; because a separate exact stage-wise token accounting
for this Dense variant is not provided, the Game-TARS token entries in
\cref{tab:training_tokens} should be interpreted as the training scale
reported for the Game-TARS model family.

At the reported stage level, OpenHA has approximately $1.30\times$ the total
token exposure of GameWAM, while the paper-reported Game-TARS recipe exceeds
the GameWAM total by more than $200\times$. These quantities provide context
for training-data scale rather than a direct measure of computational cost.
The models differ in tokenizer design, modality representation, sequence
construction, objectives, and architecture, so token exposure should not be
interpreted as FLOP-equivalent training compute.

\subsection{Cross-Modal Masking: Control, World Modeling, and Inference Efficiency}
\label{app:mask_comparison}

\cref{app:method_visibility} defines and visualizes the default
modality-decoupled mask and the joint within-block alternative in
\cref{fig:mask_decoupled,fig:mask_coupled}. We compare the two
under otherwise matched model and evaluation settings. This comparison examines
whether direct interaction between noisy video and action variables provides a
control benefit sufficient to justify the additional computation required by
joint video--action denoising.

\begin{table*}[t]
\centering
\small
\caption{
\textbf{Cross-modal masking, closed-loop performance, and online inference
efficiency.}
We report MCU Mini and MCU All ASR together with executed environment steps
averaged over successful episodes under the closed-loop protocols in
\cref{app:closed_loop_protocols}, and separately report online
execution frequency.
}
\label{tab:mask_comparison}
\setlength{\tabcolsep}{3.6pt}
\renewcommand{\arraystretch}{1.10}
\resizebox{\textwidth}{!}{
\begin{tabular}{@{}lcccccccccccc@{}}
\toprule
&
\multicolumn{3}{c}{Embodied} &
\multicolumn{3}{c}{GUI} &
\multicolumn{3}{c}{Combat} &
\multicolumn{2}{c}{Avg} &
Exec. Freq.
\\
\cmidrule(lr){2-4}
\cmidrule(lr){5-7}
\cmidrule(lr){8-10}
\cmidrule(lr){11-12}

Mask
& Steps $\downarrow$ & Mini $\uparrow$ & All $\uparrow$
& Steps $\downarrow$ & Mini $\uparrow$ & All $\uparrow$
& Steps $\downarrow$ & Mini $\uparrow$ & All $\uparrow$
& Mini $\uparrow$ & All $\uparrow$
& (Hz) $\uparrow$
\\
\midrule

Modality-decoupled (default)
    & $\mathbf{138}$
    & $\mathbf{70.0^{\pm32.2}}$
    & $\mathbf{47.5^{\pm36.0}}$
    & 155
    & $43.0^{\pm32.9}$
    & $\mathbf{60.0^{\pm38.6}}$
    & 203
    & $\mathbf{39.0^{\pm25.9}}$
    & $\mathbf{32.2^{\pm30.2}}$
    & $\mathbf{50.7}$
    & $\mathbf{46.6}$
    & $\mathbf{12.51}$
    \\

Joint video--action
    & 141
    & $\mathbf{70.0^{\pm24.1}}$
    & $40.3^{\pm35.5}$
    & $\mathbf{145}$
    & $\mathbf{45.0^{\pm34.1}}$
    & $58.8^{\pm38.5}$
    & $\mathbf{175}$
    & $24.0^{\pm25.0}$
    & $19.6^{\pm24.7}$
    & $46.3$
    & $39.6$
    & $8.12$
    \\

\bottomrule
\end{tabular}
}
\end{table*}

\cref{tab:mask_comparison} shows that the modality-decoupled design
provides stronger overall closed-loop performance. The two variants match on
Embodied Mini, while the joint model is slightly higher on GUI Mini. On the
broader MCU All evaluation, however, the default model performs better in all
three categories, with the largest difference on Combat. Overall ASR increases
from 46.3 to 50.7 on MCU Mini and from 39.6 to 46.6 on MCU All with the
modality-decoupled mask. The larger difference on MCU All suggests that the
advantage of modality decoupling becomes more apparent across the broader task
set rather than being confined to the Mini subset.

The step results provide a complementary view of interaction cost. The
modality-decoupled model requires fewer steps on Embodied tasks, whereas the
joint model reports fewer steps on GUI and Combat. These values should be read
together with task success: most notably, the lower Combat step count of the
joint model coincides with a substantial reduction in both Mini and All ASR.
The step results therefore do not change the overall pattern that the default
mask provides substantially stronger Combat control.

The computational advantage of modality decoupling is more pronounced. Joint
video--action generation operates at 8.12 Hz under the reported inference
configuration, whereas the modality-decoupled model reaches 12.51 Hz,
corresponding to approximately a $1.54\times$ higher online execution
frequency. This gain follows directly from removing iterative future-video
denoising from the online action-generation path: video prediction remains
available as a training signal, while deployment does not require the future
visual stream to be jointly denoised with the actions.

Interestingly, the optimization dynamics do not simply mirror the downstream
control results. During training, the joint model fits the action objective
faster, while optimization of the video branch progresses more slowly.
Qualitative inspection of validation predictions reveals the same asymmetry:
future-video predictions from the joint model are generally less sharp than
those produced by the modality-decoupled model. Faster fitting of the action
objective therefore does not translate into stronger overall closed-loop
control or better future-video prediction.

The magnitude of this difference also appears to depend on the visual dynamics
of the interaction regime. GUI trajectories typically contain comparatively
stable viewpoints and relatively limited frame-to-frame visual change, making
their future observations easier to predict than gameplay sequences involving
ego-motion, moving entities, and rapid scene transitions. Consistent with this
distinction, the joint variant shows little degradation on GUI: its GUI Mini
ASR is slightly higher than that of the default model (45.0 versus 43.0), while
GUI All remains close (58.8 versus 60.0). This may partly explain why joint
coupling does not produce the substantial GUI degradation observed in more
visually dynamic settings.

The difference is considerably larger on Combat, where rapid camera motion,
moving targets, and abrupt changes in scene content make the visual future more
difficult to predict. We likewise observe qualitatively that prediction blur
becomes more pronounced in fast-motion gameplay sequences, particularly for
the joint model. Together with the weaker video-side optimization, this
suggests that the consequences of cross-modal coupling may become more
significant as the visual prediction problem becomes harder.

One possible interpretation is asymmetric optimization interference between
action fitting and visual-dynamics learning. Allowing noisy future-video and
action variables to interact directly may make the action objective easier to
fit while simultaneously weakening optimization of the visual predictive
process, particularly when the visual future is intrinsically difficult to
model. This interpretation is consistent with the faster action-loss fitting,
slower video-side optimization, blurrier validation predictions, and larger
performance degradation in visually dynamic gameplay. We nevertheless treat
this as a hypothesis rather than a mechanistic conclusion, since the present
experiments do not directly isolate gradient interference or identify where
such competition arises inside the model.

This observation may also have implications beyond games. Future visual
prediction in physical embodied environments can be challenging under
ego-motion, clutter, partial observability, sensor variation, stochastic
dynamics, and other sources of visual uncertainty. Strongly coupling action
learning to an already difficult visual prediction process may therefore not
always be beneficial, and optimization pressure from action learning could
potentially interfere with learning useful visual dynamics. The
modality-decoupled formulation provides an alternative in which video
prediction continues to supply dynamics-aware training supervision without
making online action generation directly depend on uncertain future-video
variables. Whether a similar optimization asymmetry arises in physical
embodied systems remains an important direction for future investigation.

\subsection{Qualitative Closed-Loop Rollout Analysis}
\label{app:qualitative_rollouts}

Aggregate task metrics summarize whether an interaction ultimately succeeds,
but provide limited information about how the controller reaches that outcome.
We therefore visualize representative closed-loop rollouts from both Minecraft
and ViZDoom. These examples are intended as qualitative diagnostics rather than
additional benchmark results. They expose behaviors that are compressed by
aggregate success rates or episode rewards, including target search, sustained
low-level control, persistent tracking, precise GUI manipulation, recovery from
intermediate errors, survival-oriented navigation, and reactive combat.

All examples are generated by the same low-level GameWAM policy used in the
corresponding quantitative evaluations. The model receives the task instruction
and repeatedly replans from newly realized observations; the displayed
trajectories therefore reflect closed-loop execution rather than a
pre-specified open-loop action sequence.

\subsubsection{Minecraft Closed-Loop Rollouts}
\label{app:qualitative_minecraft}

\paragraph{Local exploration for target acquisition.}

\begin{figure*}[t]
    \centering
    \includegraphics[width=\textwidth]{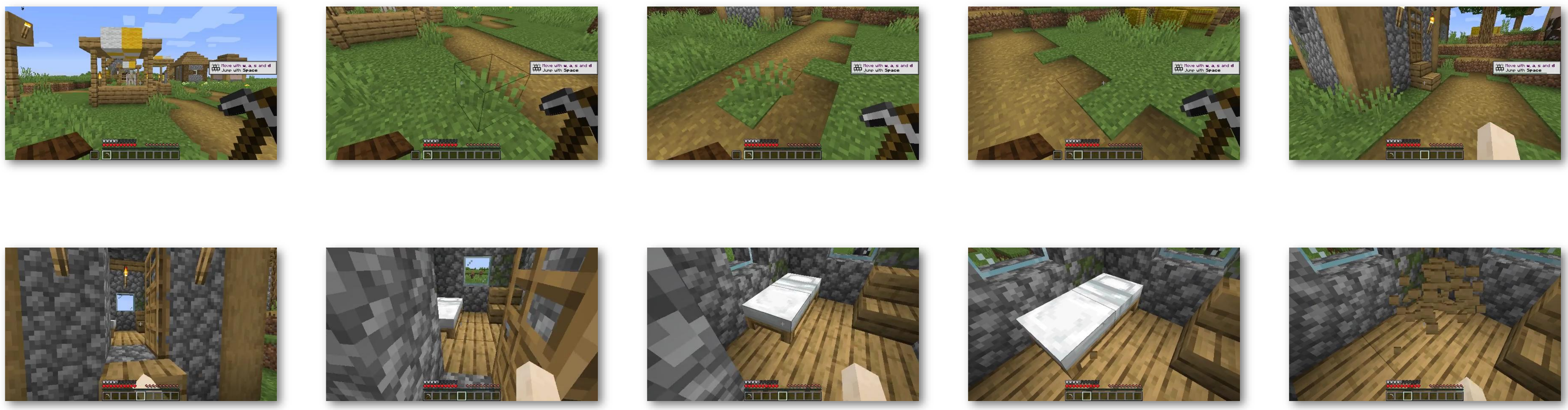}
    \caption{
    \textbf{Closed-loop rollout for \emph{``Mine the white bed.''}}
    The white bed is not visible in the initial observation. GameWAM explores the
    nearby environment, enters a structure, acquires visual access to the target,
    and completes the mining interaction. The sequence illustrates
    visually guided local exploration and target acquisition under closed-loop
    interaction.
    }
    \label{fig:rollout_white_bed}
\end{figure*}

\cref{fig:rollout_white_bed} visualizes the task
\emph{``Mine the white bed.''} The target object is not initially visible in
the agent's immediate observation. To complete the task, GameWAM first
explores the surrounding local environment through native movement and
viewpoint adjustment. The agent subsequently enters a nearby structure, where
the white bed becomes visible, and then performs the required interaction.
This rollout demonstrates that GameWAM can acquire targets that are not
directly presented in the initial view by continuously coupling action
selection with newly received visual feedback.

We interpret this behavior as a form of closed-loop local exploration and
target acquisition, rather than evidence of explicit search algorithms or
symbolic planning. GameWAM does not maintain an object-level map or a
predefined exploration policy; instead, exploratory behavior emerges from
repeated native-action prediction conditioned on evolving visual observations.

\paragraph{Maintaining a continuous interaction over an extended duration.}

\begin{figure*}[t]
    \centering
    \includegraphics[width=\textwidth]{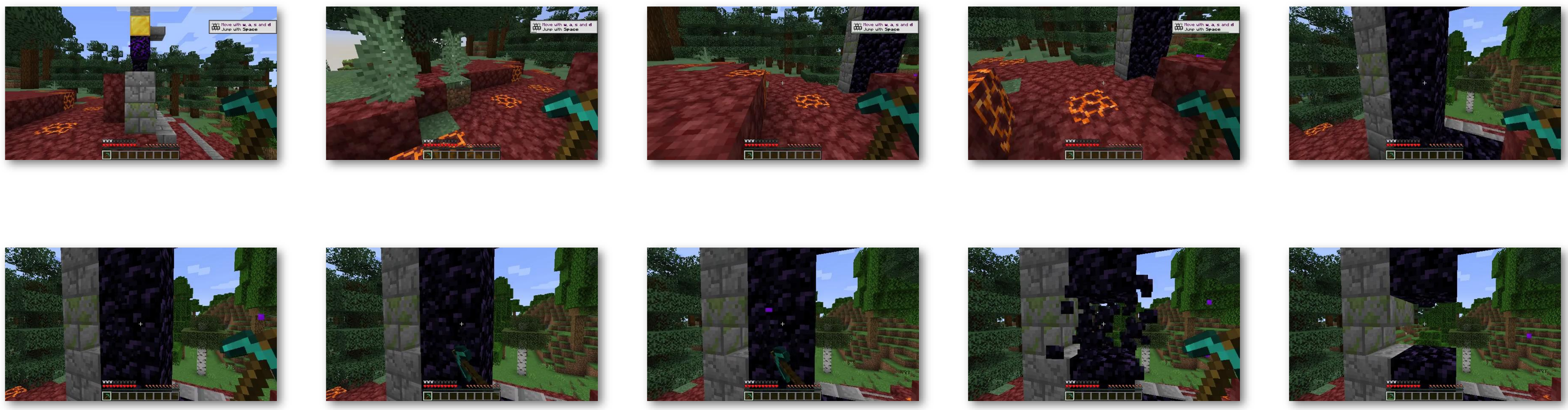}
    \caption{
    \textbf{Closed-loop rollout for
    \emph{``Collect obsidian for crafting.''}}
    Obtaining obsidian requires a sustained mining interaction rather than a
    brief action. After aligning with the target block, GameWAM maintains the
    mining behavior across successive replanning steps until the block breaks,
    illustrating temporal consistency of low-level control.
    }
    \label{fig:rollout_obsidian}
\end{figure*}

\cref{fig:rollout_obsidian} shows
\emph{``Collect obsidian for crafting.''} Mining obsidian differs from a short
attack or click interaction because the mining action must remain consistently
applied for a comparatively long interval before the block breaks. A controller
that frequently changes its action, loses alignment with the target, or
prematurely interrupts the interaction will fail to obtain the block.

GameWAM first approaches and aligns with the obsidian and then maintains the
required mining behavior until the block is successfully broken. This rollout
therefore highlights temporal consistency at the native-action level: repeated
replanning does not necessarily fragment a behavior that must remain stable
over many consecutive environment steps.

\paragraph{Long-horizon pursuit of a resilient moving target.}

\begin{figure*}[t]
    \centering
    \includegraphics[width=\textwidth]{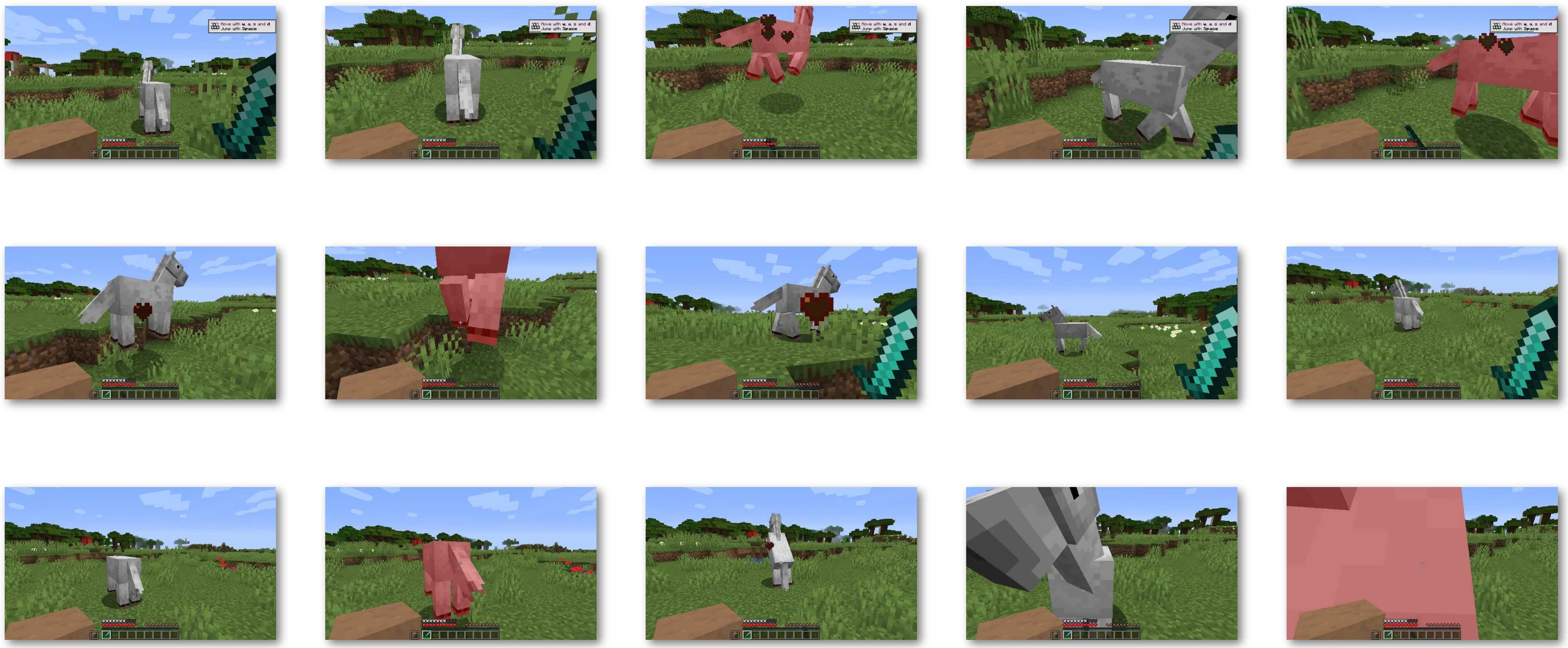}
    \caption{
    \textbf{Closed-loop rollout for \emph{``Hunt a horse.''}}
    The horse requires repeated successful attacks before task completion and
    changes its relative position throughout the interaction. GameWAM repeatedly
    reacquires the target, adjusts movement and viewpoint, and continues the
    interaction until completion. The sequence highlights persistent closed-loop
    control over an extended moving-target interaction, where consistent task
    intent must be maintained across multiple observation--action cycles.
    }
    \label{fig:rollout_horse}
\end{figure*}

\cref{fig:rollout_horse} shows the task
\emph{``Hunt a horse.''} We choose this example because horses combine two
challenging properties for low-level control. Compared with common low-health
animals such as cows, sheep, and chickens, horses require substantially more
successful attacks before task completion. Meanwhile, the target continuously
changes its relative position during the encounter. These two factors combine
to create a longer interaction horizon, requiring the controller to repeatedly
update camera orientation, movement, and attack behavior based on newly
observed visual feedback rather than relying on a short reactive sequence.

Across the rollout, GameWAM repeatedly reacquires the horse, adjusts its
position and viewpoint, and continues attacking until task completion. The
example illustrates persistent goal-directed behavior under a changing target
state: the high-level objective remains stable while the native actions needed
for completion continuously adapt throughout the interaction. This highlights
GameWAM's ability to maintain consistent behavior across multiple
observation-action cycles.

\begin{figure*}[t]
    \centering
    \includegraphics[width=\textwidth]{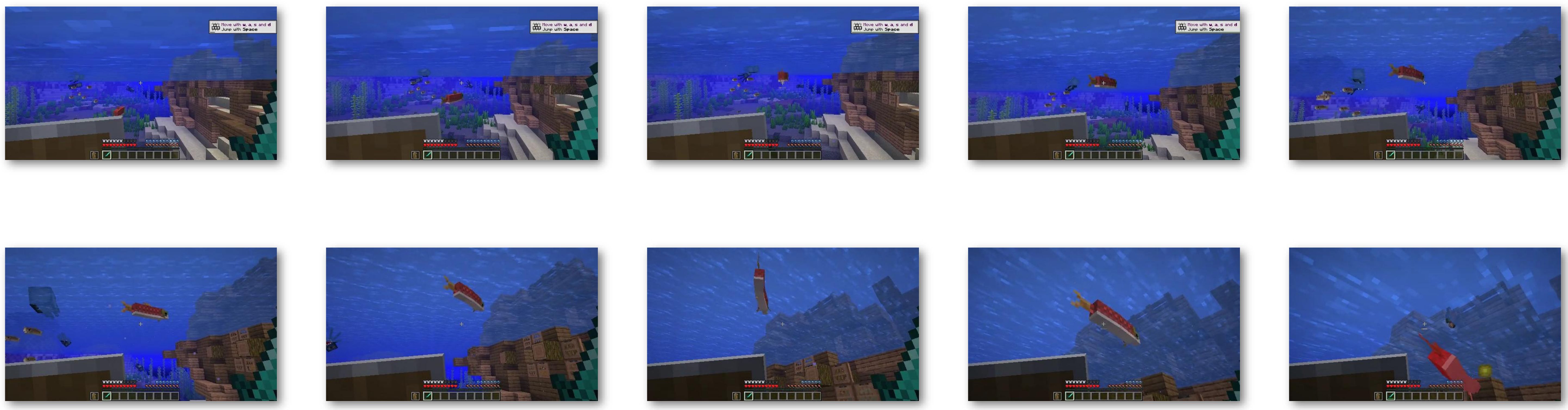}
    \caption{
    \textbf{Closed-loop rollout for \emph{``Kill the salmon.''}}
    GameWAM tracks and attacks a moving target in an underwater environment,
    where both visual appearance and locomotion differ from ordinary
    land-based gameplay. The rollout demonstrates successful closed-loop
    control under a qualitatively different gameplay condition.
    }
    \label{fig:rollout_salmon}
\end{figure*}

\paragraph{Control under underwater visual and motion dynamics.}

\cref{fig:rollout_salmon} provides a qualitatively different interaction
condition through the task \emph{``Kill the salmon.''} The entire engagement
takes place underwater. Compared with ordinary terrestrial gameplay, the scene
has different appearance, depth cues, target motion, and player locomotion.
The target also moves through three-dimensional space rather than remaining
approximately constrained to a ground plane.

GameWAM remains able to orient toward the salmon, track its changing location,
and execute the required attack sequence until completion. This example does
not by itself establish broad robustness to arbitrary visual distribution
shift, but it shows that successful native control is not restricted to the
most common land-based interaction pattern represented by the other examples.

\begin{figure*}[t]
    \centering
    \includegraphics[width=\textwidth]{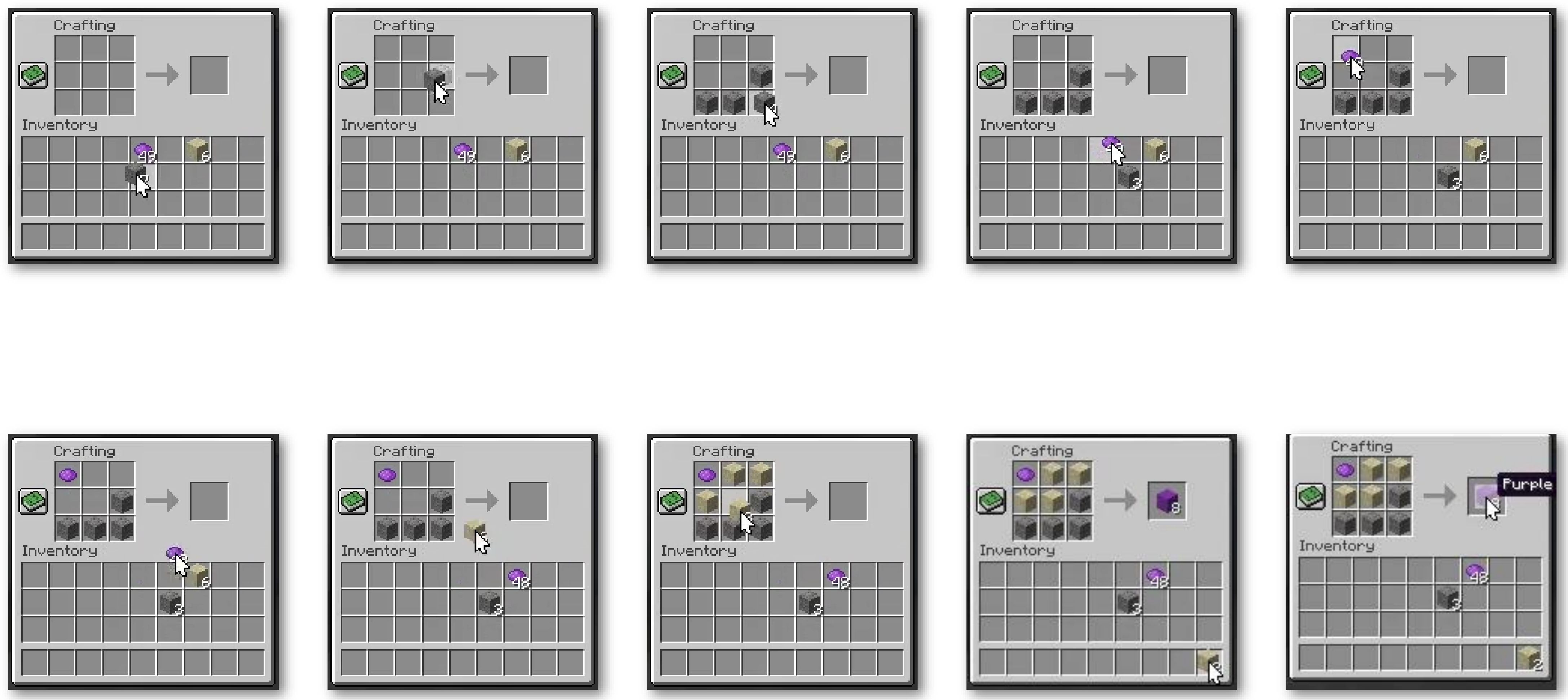}
    \caption{
    \textbf{Closed-loop rollout for
    \emph{``craft item purple concrete powder.''}}
    GameWAM progressively moves the required ingredients from the inventory
    into their corresponding crafting-grid locations and completes the target
    recipe. The sequence illustrates precise cursor control and structured
    multi-step placement through the native GUI action interface.
    }
    \label{fig:rollout_purple}
\end{figure*}

\begin{figure*}[t]
    \centering
    \includegraphics[width=\textwidth]{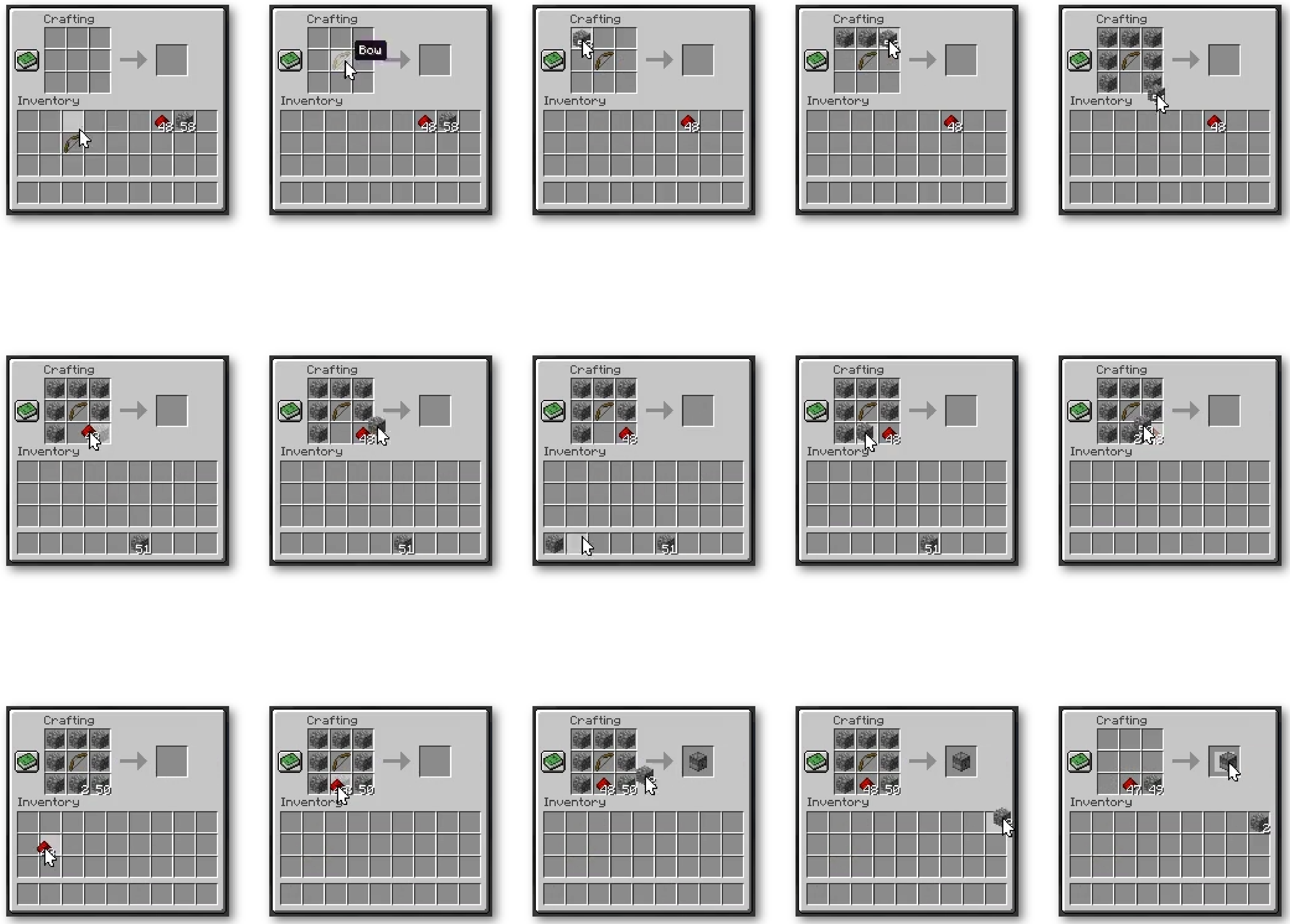}
    \caption{
    \textbf{Closed-loop rollout for
    \emph{``Create a dispenser in the crafting table.''}}
    GameWAM makes an intermediate placement error while constructing the
    recipe. After observing the resulting GUI state, subsequent actions adjust
    the crafting-grid configuration and complete the target item. The rollout
    illustrates error recovery enabled by repeated closed-loop observation and
    replanning.
    }
    \label{fig:rollout_dispenser}
\end{figure*}

\paragraph{Precise spatial placement in the crafting grid.}

Gameplay and GUI interaction place substantially different demands on the same
native keyboard--mouse interface. Gameplay primarily requires camera-centered
movement and interaction, whereas crafting requires the cursor to manipulate
small spatial targets under recipe-specific placement constraints.

\cref{fig:rollout_purple} shows
\emph{``craft item purple concrete powder.''} Successful execution requires
multiple ingredient types to be moved from the inventory into specific
locations in the crafting grid. The task therefore depends not only on
selecting the correct objects, but also on controlling their relative
positions and completing a multi-step placement sequence through continuous
cursor motion and discrete mouse actions.

The rollout shows GameWAM progressively constructing the required arrangement
and obtaining the target output. This example highlights the role of
mode-specific GUI action modeling: the same physical mouse channels used for
camera control during gameplay must here support precise cursor displacement,
selection, and placement inside a structured interface.

\paragraph{Recovery from an intermediate placement error.}

\cref{fig:rollout_dispenser} shows
\emph{``Create a dispenser in the crafting table.''} This trajectory is
particularly informative because successful execution is not obtained through
a perfectly correct initial sequence. During crafting, GameWAM makes an
intermediate placement error that leaves the grid in an incorrect state.
Subsequent observations expose that state to the controller, after which the
policy modifies the placement and eventually constructs the valid recipe.

The example illustrates a basic but important consequence of closed-loop GUI
control. An intermediate low-level error need not irreversibly determine the
remainder of the trajectory: because the model replans from the realized
interface state, later actions can compensate for an earlier mistake. We refer
to this behavior as closed-loop error recovery rather than high-level
self-reflection, since no explicit symbolic diagnosis or textual reasoning
process is assumed.

\subsubsection{ViZDoom Closed-Loop Rollouts}
\label{app:qualitative_vizdoom}

The ViZDoom examples complement Minecraft by examining GameWAM under faster,
more reactive combat dynamics. Unlike the Minecraft cases above, these tasks
primarily reward continuous survival and combat effectiveness rather than
completion of a discrete object-manipulation goal. The four maps also impose
different spatial structures: the two Battle scenarios require locomotion and
target engagement through navigable environments, whereas the two defense
scenarios emphasize rapid target selection under constrained movement.

\paragraph{Reactive combat and survival in Battle 1.}

\begin{figure*}[t]
    \centering
    \includegraphics[width=\textwidth]{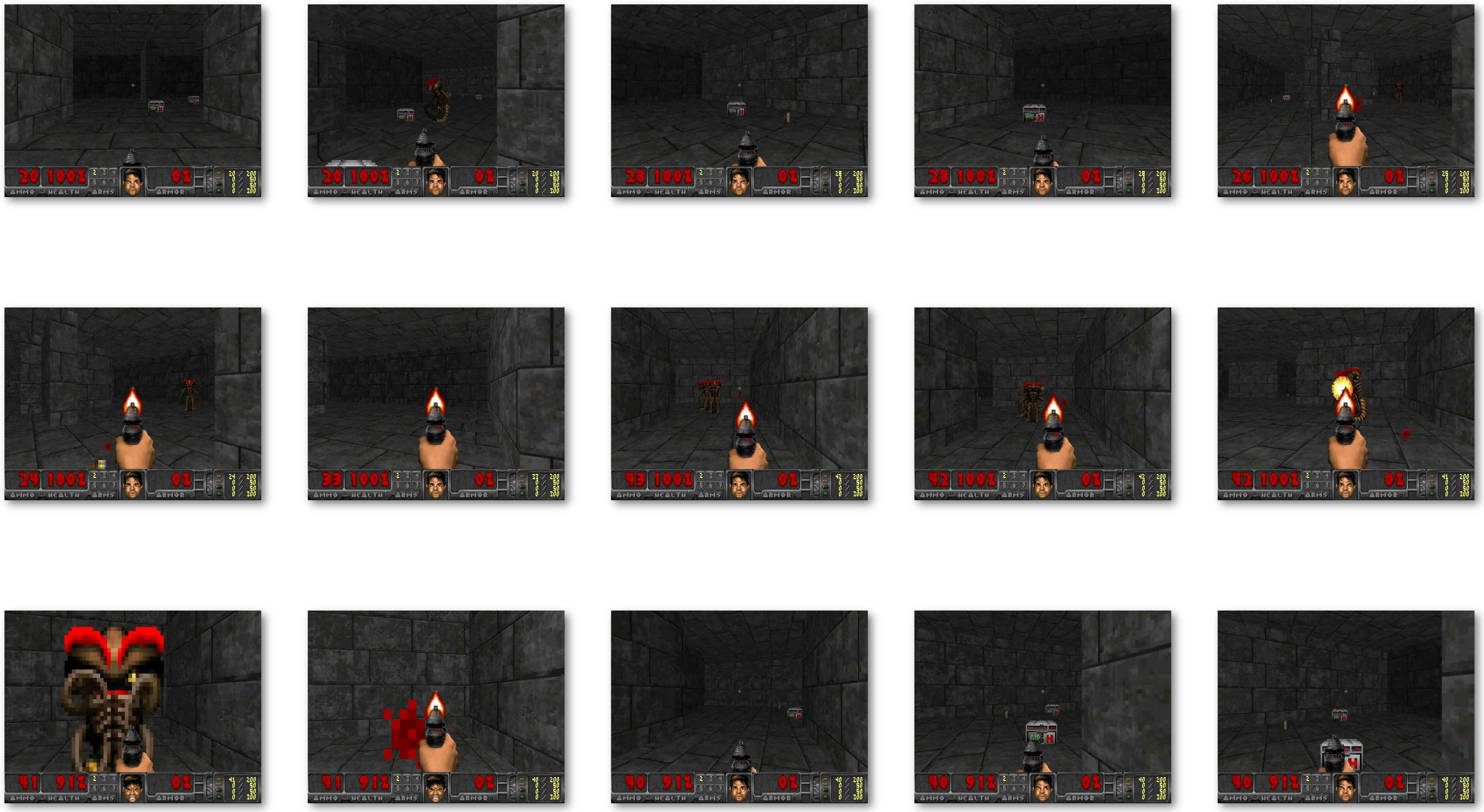}
    \caption{
    \textbf{Closed-loop rollout on ViZDoom Battle 1 for
    \emph{``Fight off the monsters and stay alive as long as you can.''}}
    GameWAM navigates through the environment, reacts to newly visible enemies,
    adjusts its view toward threats, and repeatedly engages them while
    continuing the survival-oriented rollout.
    }
    \label{fig:rollout_vizdoom_battle1}
\end{figure*}

\cref{fig:rollout_vizdoom_battle1} visualizes the instruction
\emph{``Fight off the monsters and stay alive as long as you can.''}
The rollout alternates between navigation and combat as enemies enter the
agent's field of view. GameWAM changes orientation to bring nearby threats into
the firing direction and engages them while continuing to move through the
environment.

This task differs from a single-target Minecraft combat interaction because
there is no terminal target whose elimination completes the instruction.
Instead, the controller must repeatedly react to newly encountered threats
while preserving its ability to continue the episode. The sequence therefore
illustrates sustained closed-loop combat in which movement, target acquisition,
and firing are repeatedly recomputed from new observations.

\begin{figure*}[t]
    \centering
    \includegraphics[width=\textwidth]{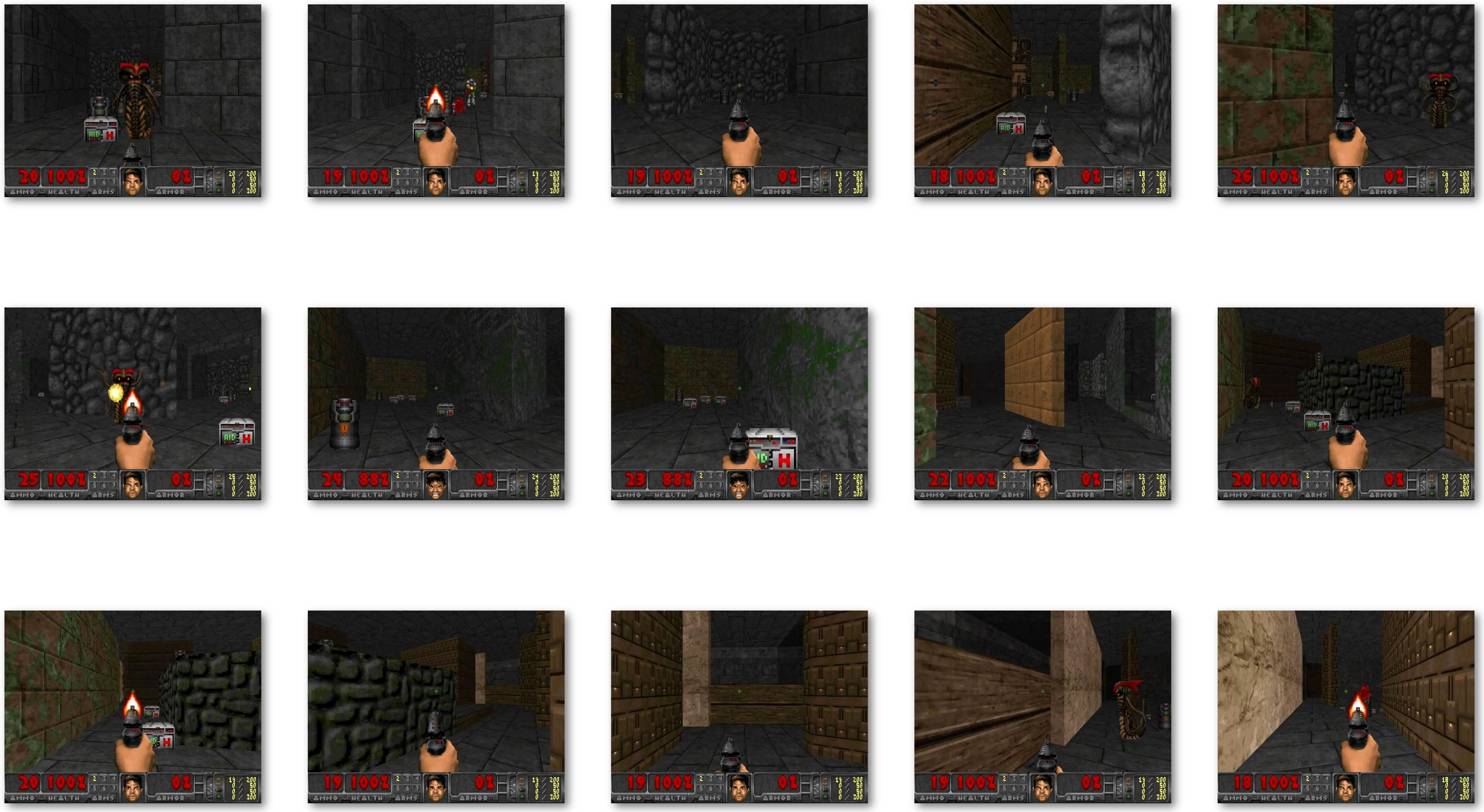}
    \caption{
    \textbf{Closed-loop combat and health recovery on ViZDoom Battle 2 for
    \emph{``Fight off the monsters in the maze and stay alive as long as you
    can.''}}
    GameWAM navigates and fights through the maze. After taking damage, the
    policy approaches and collects a health pack, recovers health, and
    subsequently continues the survival-oriented combat rollout. The sequence
    illustrates adaptation to realized agent state in addition to reactive
    enemy engagement.
    }
    \label{fig:rollout_vizdoom_battle2}
\end{figure*}

\paragraph{Combat with health-aware recovery in Battle 2.}

\cref{fig:rollout_vizdoom_battle2} shows the more spatially complex
instruction
\emph{``Fight off the monsters in the maze and stay alive as long as you
can.''}
Here the agent must combine combat with navigation through a maze-like
environment, where walls and corridors repeatedly change which enemies and
resources are visible.

The rollout also exposes an important form of closed-loop adaptation that is
not captured by the final episode reward alone. After taking damage during an
engagement, GameWAM does not simply continue attacking until failure. It
navigates toward an available health pack, collects it to recover health, and
then continues through the environment and resumes combat. The sequence
therefore demonstrates that the policy can condition its subsequent behavior
on the realized consequences of earlier interaction, including changes in its
own state.

We interpret this behavior as state-sensitive closed-loop recovery rather than
explicit resource planning. No symbolic survival planner is provided; the
change in behavior occurs through repeated prediction from the updated visual
and proprioceptive state.

\paragraph{Maintaining a frontal defense line.}

\begin{figure*}[t]
    \centering
    \includegraphics[width=\textwidth]{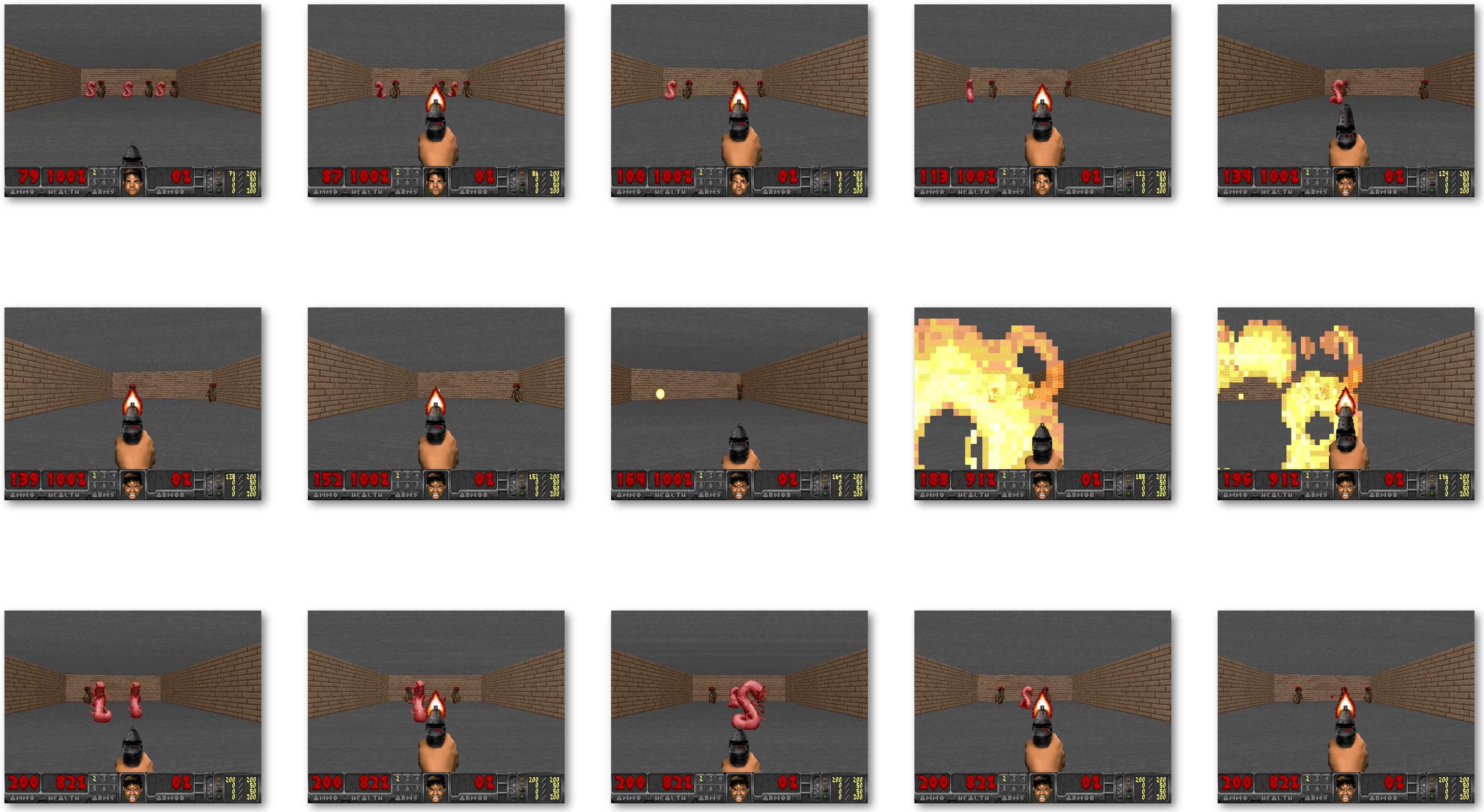}
    \caption{
    \textbf{Closed-loop rollout for
    \emph{``Defend the line: shoot the monsters advancing toward you from
    across the room.''}}
    GameWAM repeatedly acquires approaching enemies, adjusts horizontal aim,
    and fires as the active threat changes. The sequence highlights reactive
    closed-loop control under a frontal-defense objective.
    }
    \label{fig:rollout_vizdoom_defend_line}
\end{figure*}

\cref{fig:rollout_vizdoom_defend_line} visualizes
\emph{``Defend the line: shoot the monsters advancing toward you from across
the room.''}
Unlike the Battle maps, the spatial objective is comparatively constrained:
enemies approach primarily from the forward region, and successful behavior
requires repeated target acquisition and firing as successive threats advance.

GameWAM continually adjusts its horizontal aim between approaching enemies and
engages targets before they reach the defended position. The rollout
illustrates rapid visual--action feedback in a setting where the relevant
target can change from one replanning step to the next and delayed correction
would allow enemies to continue advancing.

\paragraph{Multi-directional threat response in Defend the Center.}

\begin{figure*}[t]
    \centering
    \includegraphics[width=\textwidth]{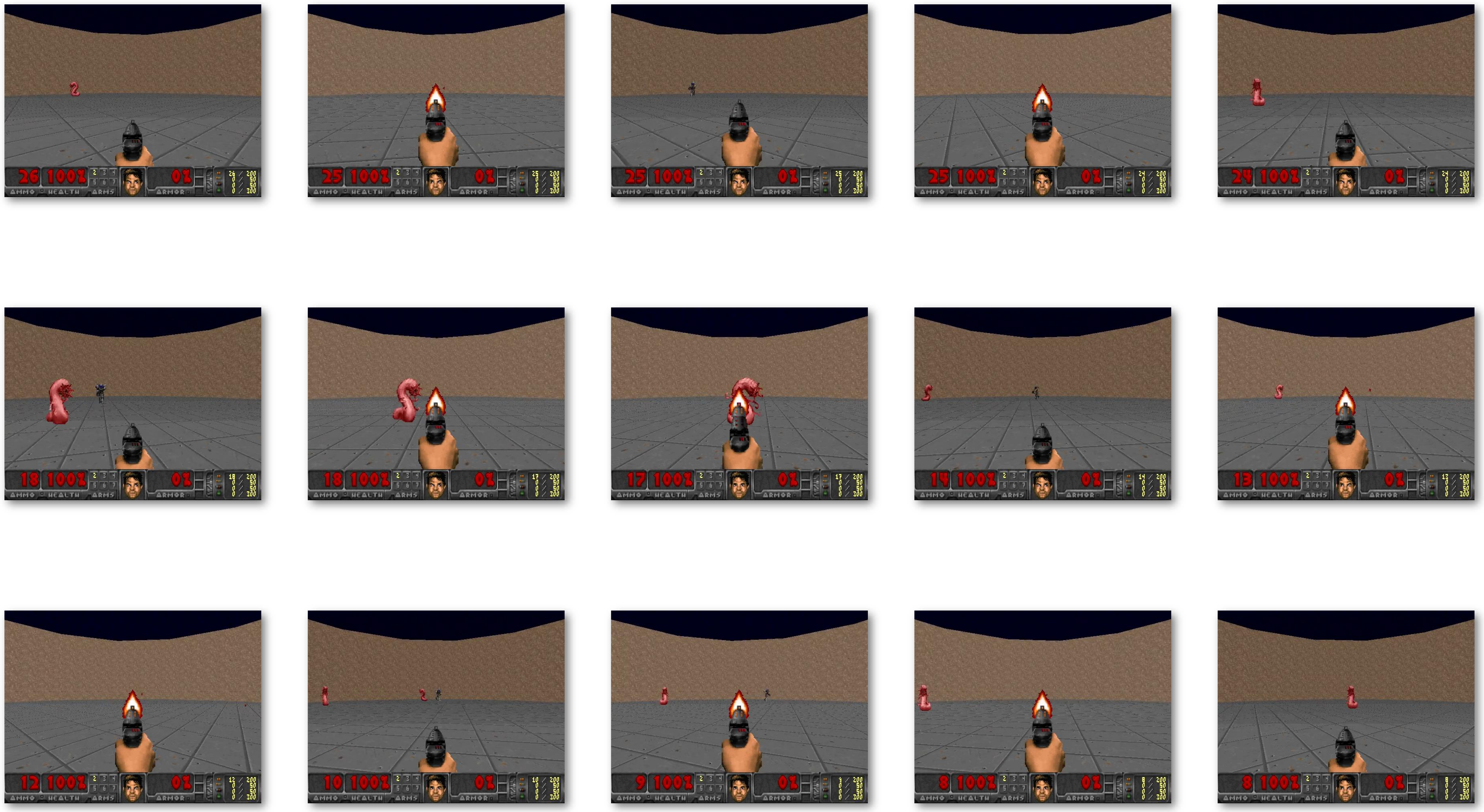}
    \caption{
    \textbf{Closed-loop rollout for
    \emph{``Defend the center: shoot the enemies closing in from all
    directions.''}}
    Threats can appear from different directions around the agent. GameWAM
    repeatedly redirects its view and switches targets as enemies approach,
    illustrating reactive multi-directional defense from a central position.
    }
    \label{fig:rollout_vizdoom_defend_center}
\end{figure*}

\cref{fig:rollout_vizdoom_defend_center} considers
\emph{``Defend the center: shoot the enemies closing in from all directions.''}
The control challenge differs from Defend the Line because threats are not
restricted to a single frontal direction. The policy must repeatedly change
view orientation to locate and engage enemies appearing around the central
position.

Across the rollout, GameWAM redirects its aim between different parts of the
arena and fires on enemies as they enter actionable views. This example
therefore emphasizes repeated orientation changes and target switching rather
than pursuit through the environment. Together with Defend the Line, it shows
that the same learned controller can realize qualitatively different combat
patterns under different language-conditioned objectives.

\paragraph{Qualitative interpretation.}

The Minecraft and ViZDoom rollouts expose complementary properties of the same
world--action policy. Minecraft emphasizes longer-horizon interaction:
searching for an initially non-visible object, maintaining a long mining
action, pursuing moving entities, manipulating structured interfaces, and
recovering from an intermediate GUI error. ViZDoom instead emphasizes rapid
feedback under continuously changing combat states: repeated target
acquisition, survival-oriented navigation, adaptation after damage, and
switching between threats under different spatial constraints.

These examples also illustrate why closed-loop replanning is important beyond
aggregate task success. Useful behavior need not correspond to a single
unchanging action pattern. Obsidian mining requires maintaining an action
despite repeated replanning, horse hunting requires preserving task intent
while continually changing movement and attack commands, the dispenser example
requires correcting an earlier interaction, and Battle 2 requires changing
behavior after the agent's health state changes. In each case, subsequent
actions are conditioned on realized environment feedback rather than on an
unexecuted prediction of what should have happened.

As with any qualitative visualization, these trajectories should not be
interpreted as estimates of how frequently each behavior occurs. They are
selected rollouts intended to expose interaction patterns that are difficult
to convey through aggregate success rate or reward alone. Quantitative
Minecraft and ViZDoom performance remains reported through the evaluation
results in the main paper.

\subsection{Cross-Game Zero-Shot Transfer to VoxeLibre}
\label{app:voxelibre_transfer}

We further probe cross-game transfer by directly deploying the
Minecraft-trained GameWAM checkpoint in VoxeLibre without any
environment-specific training or fine-tuning. The final checkpoint after the
second Minecraft training epoch is used unchanged. Model-side configurations,
including the $224\times224$ observation resolution and temporal settings,
remain identical to the Minecraft evaluation. We only apply a deterministic
interface mapping between VoxeLibre keyboard--mouse controls and the native
action representation expected by GameWAM.

Because the GUI conventions of the two games differ substantially, this
diagnostic focuses on basic gameplay interactions rather than interface tasks.
We define six simple tasks spanning resource collection, combat, and block
placement and provide each through a short natural-language instruction
(e.g., ``Chop down the oak log.''). Each task is evaluated over 20 episodes.
The primary metric is success rate; average executed environment steps are
reported over successful episodes.

\begin{table}[t]
\centering
\small
\caption{
\textbf{Zero-shot cross-game transfer from Minecraft to VoxeLibre.}
The Minecraft-trained GameWAM checkpoint is evaluated without VoxeLibre
training or fine-tuning. Each task is evaluated over 20 episodes; steps are
averaged over successful episodes.
}
\label{tab:voxelibre_zero_shot}
\setlength{\tabcolsep}{7.0pt}
\renewcommand{\arraystretch}{1.2}
\begin{tabular}{@{}lccc@{}}
\toprule
Task & Success & Rate & Steps / success \\
\midrule
Chop Tree
    & 17/20 & \textbf{85.0\%} & 61.8 \\
Mine Stone
    & 14/20 & 70.0\% & 41.9 \\
Mine Iron Ore
    & 10/20 & 50.0\% & 49.4 \\
Kill Cow
    & 2/20 & 10.0\% & 46.5 \\
Kill Zombie
    & 12/20 & 60.0\% & 112.6 \\
Place Block
    & 16/20 & 80.0\% & 67.9 \\
\midrule
\textbf{Overall}
    & \textbf{71/120} & \textbf{59.2\%} & -- \\
\bottomrule
\end{tabular}
\end{table}

As shown in \cref{tab:voxelibre_zero_shot}, GameWAM succeeds in 71 of 120 episodes, corresponding to a 59.2\% micro-averaged success rate. Success rates vary substantially across tasks, ranging from 10.0\% on Kill Cow to 85.0\% on Chop Tree, with strong transfer also observed on Place Block (80.0\%) and Mine Stone (70.0\%). These results suggest that part of the learned visual--action control behavior transfers to a related but distinct game despite changes in rendering and interaction dynamics.

This experiment is intended as a diagnostic of cross-game generalization
rather than as a new benchmark or a claim of broad out-of-domain transfer.
Minecraft and VoxeLibre share related voxel-world structure and basic gameplay
semantics, and the evaluated tasks deliberately focus on low-level gameplay
skills for which a deterministic native-control interface mapping is possible.

\subsection{Extended LASI Analysis}
\label{app:lasi}

We use the LASI notation, source variable, and DCT convention defined in
\cref{sec:lasi}. This section provides additional source-space
controls, frequency-resolved analyses, generative-trajectory diagnostics, and
closed-loop measurements underlying the main-text findings.

\subsubsection{Evaluation Setup and Frequency Decomposition}

The qualitative LASI failure pattern was observed across multiple GameWAM
checkpoints spanning different training stages, hyperparameter settings, and
training configurations. To keep the controlled intervention suite internally
consistent, all quantitative diagnostics reported below are performed on the
final checkpoint after the second Minecraft training epoch, used here as the
representative checkpoint, so every comparison uses the same underlying
policy. Controlled source interventions hold the conditioning context fixed
and vary only the explicit action source. For analyses that resolve source
sensitivity along the generative trajectory, we use a deterministic 20-step
integration path, providing finer denoising-time resolution than the standard
closed-loop sampling configuration. This diagnostic setting is used only for
the LASI analysis and does not alter the reported task-performance evaluation.

The controlled explicit-source matrix contains 24 fixed conditions and
16 action sources per condition ($24\times16=384$ samples). Low-frequency
donor swapping and low-frequency zeroing each use the same controlled matrix;
the antithetic analysis uses $24\times8=192$ source pairs. Confidence
intervals reported below use 10,000 condition-cluster bootstrap resamples
where available.

For each analyzed eight-step continuous camera chunk, we apply an orthonormal
DCT along time. Modes $q=0,1,2$ form the low-frequency band and modes
$q=3,\ldots,7$ form the high-frequency band. P0--P2 and Y0--Y2 denote pitch
and yaw DCT modes 0--2, respectively.

\subsubsection{Within-Condition Source Association}

For condition $c$, source $j$, action dimension $d$, and DCT mode $q$, define
\begin{equation}
    r_{c,d,q}
    =
    \operatorname{Corr}_j
    \left(
      \widetilde Z_{c,j,d,q},
      \widehat{\widetilde X}_{0,c,j,d,q}
    \right).
\end{equation}

The six low-frequency modes show substantial within-condition association.
Averaging $r_{c,d,q}$ across the fixed conditions, pitch DCT0--2 have mean
correlations $0.776$, $0.613$, and $0.746$, while yaw DCT0--2 have
$0.890$, $0.824$, and $0.746$. The yaw DCT0 95\% confidence interval is
$[0.835,0.936]$. Because the condition is fixed within each group, this
analysis isolates source-associated variation, but correlation alone is not
used as the causal claim.

\subsubsection{Donor, Zeroing, and Source-Space Controls}

For a base source $Z^b$ and donor $Z^d$, the low-frequency intervention is
\begin{equation}
    \widetilde Z_q^{\mathrm{swap}}
    =
    \begin{cases}
      \widetilde Z_q^d, & q\in\{0,1,2\},\\
      \widetilde Z_q^b, & q\in\{3,\ldots,7\}.
    \end{cases}
\end{equation}

The donor-follow rate asks whether the intervened output moves closer to the
corresponding donor output than to the base output. For yaw DCT0, the
donor-follow rate is 94.8\% (95\% CI $[92.2,97.1]\%$), and the correlation
between donor-induced source and output changes is $r=0.921$.

For low-frequency zeroing, $\widetilde Z_q=0$ for
$q\in\{0,1,2\}$. Writing $N_c$ and $N_s$ for the numbers of fixed conditions
and sources, respectively, define the condition-averaged within-condition
source variances
\begin{equation}
V_{d,q}^{\mathrm{base}}
=
\frac{1}{N_c}
\sum_{c=1}^{N_c}
\operatorname{Var}_j
\left[
\widehat{\widetilde X}^{\mathrm{base}}_{0,c,j,d,q}
\right],
\qquad
V_{d,q}^{\mathrm{zero}}
=
\frac{1}{N_c}
\sum_{c=1}^{N_c}
\operatorname{Var}_j
\left[
\widehat{\widetilde X}^{\mathrm{zero}}_{0,c,j,d,q}
\right].
\end{equation}
The fraction of source-induced variance removed is
\begin{equation}
R_{d,q}^{\mathrm{zero}}
=
1-
\frac{V_{d,q}^{\mathrm{zero}}}
     {V_{d,q}^{\mathrm{base}}}.
\end{equation}

Yaw DCT0 loses 99.25\% of its source-induced variance
(95\% CI $[98.93,99.50]\%$). The corresponding reductions for pitch DCT0--2
are 95.95\%, 91.44\%, and 93.21\%; for yaw DCT1--2 they are 98.45\% and
96.99\%.

To compare variation across conditions with variation induced by the source,
we additionally define
\begin{equation}
R_{d,q}^{\mathrm{cond/src}}
=
\frac{
\operatorname{Var}_c
\left[
\frac{1}{N_s}
\sum_{j=1}^{N_s}
\widehat{\widetilde X}^{\mathrm{base}}_{0,c,j,d,q}
\right]
}{
V_{d,q}^{\mathrm{base}}
}.
\end{equation}

We also use antithetic source pairs $Z$ and $-Z$:
\begin{equation}
    \widehat X_{\mathrm{odd}}(Z)
    =\tfrac12[\widehat X(Z)-\widehat X(-Z)],
    \qquad
    \widehat X_{\mathrm{even}}(Z)
    =\tfrac12[\widehat X(Z)+\widehat X(-Z)].
\end{equation}
Let $\mathcal J_{\pm}$ denote the subset of base sources paired with their
negatives. The antithetic cancellation ratio for the corresponding DCT
coefficient is
\begin{equation}
R_{d,q}^{\mathrm{anti}}
=
1-
\frac{
\frac{1}{N_c}
\sum_c
\operatorname{Var}_{j\in\mathcal J_{\pm}}
\left[
\widehat{\widetilde X}_{\mathrm{even},c,j,d,q}
\right]
}{
\frac{1}{N_c}
\sum_c
\operatorname{Var}_{j\in\mathcal J_{\pm}}
\left[
\widehat{\widetilde X}^{\mathrm{base}}_{0,c,j,d,q}
\right]
}.
\end{equation}

\begin{figure*}[t]
    \centering
    \includegraphics[width=\textwidth]
    {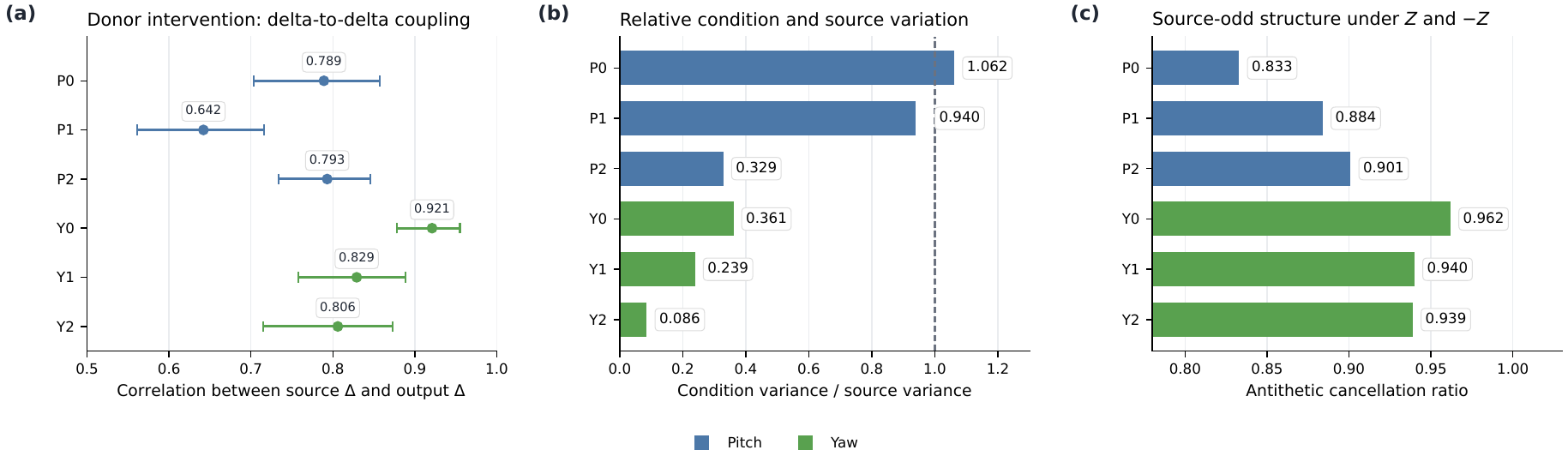}
    \caption{
    \textbf{LASI source-space controls.}
    \textbf{(a)} Correlation between the change imposed on a low-frequency
    source coefficient and the corresponding change in the predicted
    clean-action coefficient.
    \textbf{(b)} Condition-induced variance divided by source-induced variance;
    values below one indicate larger source variation for the measured mode.
    \textbf{(c)} Antithetic cancellation ratio for paired $Z$ and $-Z$ sources,
    measuring source-odd structure. P0--P2 and Y0--Y2 denote pitch and yaw
    DCT modes 0--2.
    }
    \label{fig:app_source_diagnostics}
\end{figure*}

\cref{fig:app_source_diagnostics} summarizes the source-space controls.
The condition/source variance ratio should not be interpreted as source
dominance in every mode: pitch DCT0 is 1.062, slightly above one, whereas the
remaining reported ratios are 0.940, 0.329, 0.361, 0.239, and 0.086.

\subsubsection{Frequency Selectivity}

We compare three unit-norm perturbation directions: a low-frequency direction
aligned with the leading principal direction of clean camera actions in DCT
modes 0--2, an orthogonal direction within the same low-frequency subspace,
and a high-frequency direction in modes 3--7. Thus, differences in response
are not attributable to perturbation magnitude.

For a fixed condition $c$ and perturbation direction $q$, write
$\widehat X_0(x;\sigma,c)$ for the reconstructed clean-action estimate
obtained from noisy action state $x$ at noise level $\sigma$. We estimate the
local response by the centered finite difference
\begin{equation}
    J_{\sigma,c}(q)
    \approx
    \frac{
      \widehat X_0
      (X_{\sigma,c}+\epsilon q;\sigma,c)
      -
      \widehat X_0
      (X_{\sigma,c}-\epsilon q;\sigma,c)
    }{2\epsilon}.
\end{equation}
We measure the induced low-frequency response magnitude as
\begin{equation}
    \rho_{\sigma,c}(q)
    =
    \left\|
      P_{\mathrm{low}}J_{\sigma,c}(q)
    \right\|_2,
\end{equation}
where $P_{\mathrm{low}}$ projects the continuous camera trajectory onto DCT
modes 0--2. The reported comparison uses $\epsilon=0.05$.

The primary frequency-selectivity statistic compares the aligned
low-frequency perturbation with the high-frequency perturbation through the
condition-paired log-response difference
\begin{equation}
\Delta_\sigma^{\mathrm{low/high}}
=
\frac{1}{N_c}
\sum_{c=1}^{N_c}
\left[
\log\rho_{\sigma,c}(q_{\mathrm{low}})
-
\log\rho_{\sigma,c}(q_{\mathrm{high}})
\right],
\qquad
G_\sigma^{\mathrm{low/high}}
=
\exp\!\left(
\Delta_\sigma^{\mathrm{low/high}}
\right).
\end{equation}

Across the tested
$\sigma\in\{0.20,0.40,0.55,0.68,0.80\}$, the geometric low/high response
ratios are 4.35, 4.20, 3.53, 4.58, and 4.43. The aligned low-frequency
direction is also compared with the orthogonal low-frequency direction; this
difference is substantially smaller and depends on noise level. The supported
conclusion is therefore strong selectivity between the tested low- and
high-frequency perturbations, while evidence for additional selectivity among
directions \emph{within} the low-frequency subspace is weaker and
noise-dependent.

\subsubsection{Single-Forward Source Transfer}

To separate one-call source transfer from iterative path effects, we directly
construct
\begin{equation}
    X_\sigma^{\mathrm{ana}}
    =(1-\sigma)X_0+\sigma Z
\end{equation}
and evaluate the denoiser once at that fixed noise level, without taking an
integration step.

To remove between-condition variation, source transfer is measured from
within-condition source-pair differences. For two sources $j$ and $k$ under
the same condition,
\begin{equation}
\Delta\widetilde Z_{c,jk,d,q}
=
\widetilde Z_{c,k,d,q}
-
\widetilde Z_{c,j,d,q},
\qquad
\Delta\widehat{\widetilde X}_{0,c,jk,d,q}^{\mathrm{ana}}
=
\widehat{\widetilde X}_{0,c,k,d,q}^{\mathrm{ana}}
-
\widehat{\widetilde X}_{0,c,j,d,q}^{\mathrm{ana}}.
\end{equation}
The source-to-output gain for a DCT coefficient is the regression slope
\begin{equation}
    g_{\mathrm{ana}}(\sigma)
    =
    \frac{
      \operatorname{Cov}_{c,j<k}
      \left(
        \Delta\widetilde Z_{c,jk,d,q},
        \Delta\widehat{\widetilde X}_{0,c,jk,d,q}^{\mathrm{ana}}
      \right)
    }{
      \operatorname{Var}_{c,j<k}
      \left(
        \Delta\widetilde Z_{c,jk,d,q}
      \right)
    }.
\end{equation}

\begin{figure*}[t]
    \centering
    \includegraphics[width=\textwidth]
    {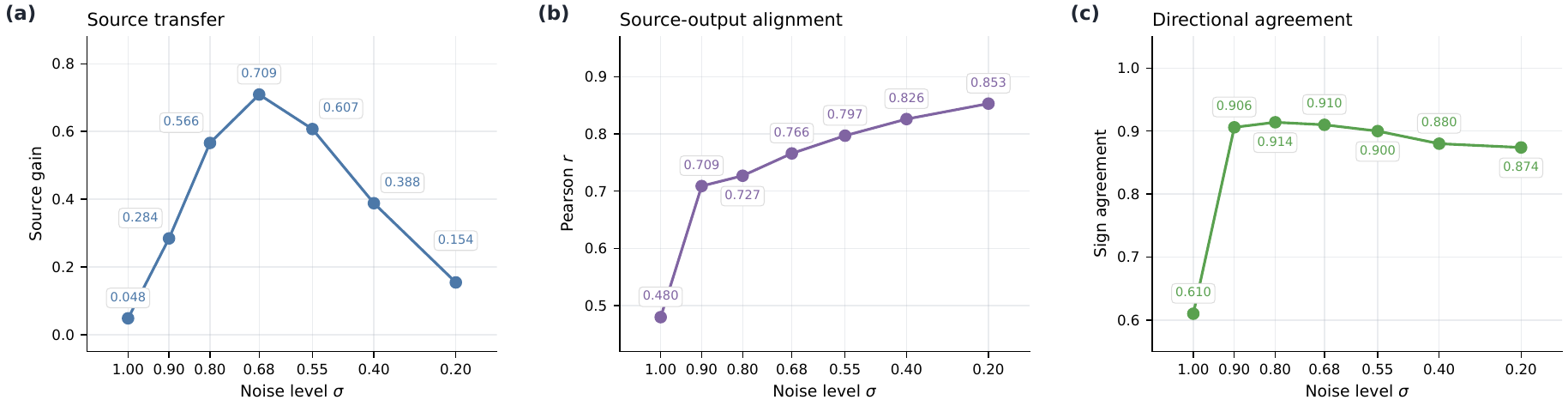}
    \caption{
    \textbf{Single-forward LASI across analytic noise levels.}
    \textbf{(a)} Regression gain from within-condition yaw-DCT0 source
    differences to the corresponding reconstructed clean-action differences.
    \textbf{(b)} Pearson correlation between the same source and reconstructed
    action differences.
    \textbf{(c)} Sign agreement of the paired differences. Each point uses
    one denoiser evaluation at an analytically constructed noisy state, with
    no integration update before measurement.
    }
    \label{fig:app_single_forward}
\end{figure*}

\cref{fig:app_single_forward} shows that source transfer is already
present before iterative denoising. For yaw DCT0, the gain peaks at 0.709 at
$\sigma=0.68$ in this sweep; the corresponding pairwise Pearson correlation
increases from 0.480 at $\sigma=1$ to 0.853 at $\sigma=0.2$, while sign
agreement is approximately 0.87--0.91 over most measured noise levels after
the initial point.

\subsubsection{Model-Path Amplification}
\label{app:lasi_path}

For a state reached by the iterative model path at the same numeric noise
level, define $g_{\mathrm{path}}(\sigma)$ analogously using within-condition
source-pair differences. The absolute and relative amplification statistics
are
\begin{equation}
    A(\sigma)
    =
    g_{\mathrm{path}}(\sigma)
    -
    g_{\mathrm{ana}}(\sigma),
    \qquad
    R_g(\sigma)
    =
    \frac{g_{\mathrm{path}}(\sigma)}
         {|g_{\mathrm{ana}}(\sigma)|}.
\end{equation}

\begin{figure*}[t]
    \centering
    \includegraphics[width=\textwidth]
    {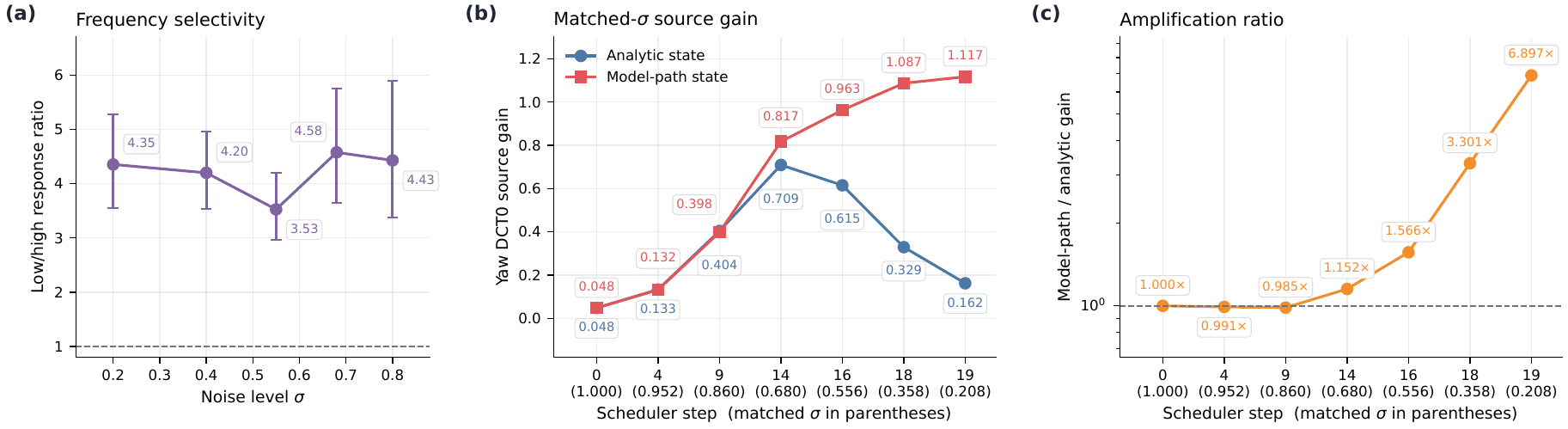}
    \caption{
    \textbf{LASI frequency selectivity and model-path amplification.}
    \textbf{(a)} Geometric response ratio of aligned low-frequency versus
    high-frequency perturbations over tested noise levels.
    \textbf{(b)} Yaw-DCT0 source gain for an analytically constructed noisy
    state and for the state reached along the iterative model path at the
    matched $\sigma$ shown in parentheses.
    \textbf{(c)} Model-path gain divided by the magnitude of the analytic gain;
    the endpoint ratio reaches 6.897$\times$ at step 19
    ($\sigma=0.208$).
    }
    \label{fig:app_lasi_frequency}
\end{figure*}

\cref{fig:app_lasi_frequency} combines the frequency-selectivity result
with the matched-$\sigma$ analytic/model-path comparison. At early denoising
steps, the two gains are nearly identical. A positive gap appears in the
mid/low-noise region; at step 19 ($\sigma=0.208$), yaw DCT0 has analytic gain
0.162 and model-path gain 1.117, giving a 6.897$\times$ ratio.

\begin{figure}[t]
    \centering
    \includegraphics[width=0.92\linewidth]
    {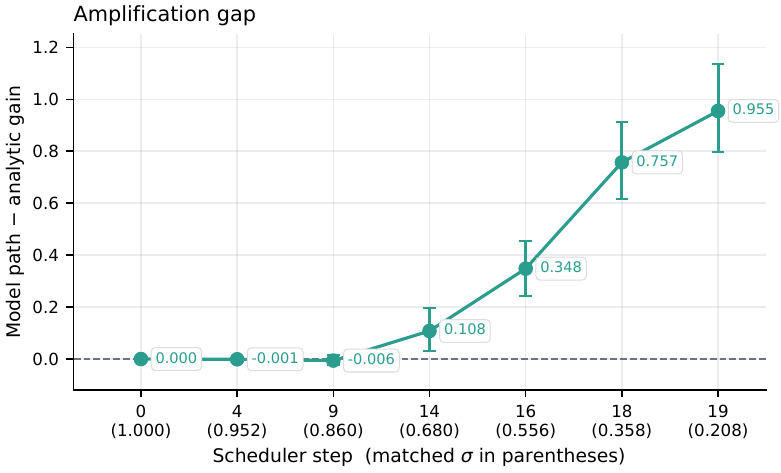}
    \caption{
    \textbf{LASI model-path amplification gap.}
    Difference $g_{\mathrm{path}}-g_{\mathrm{ana}}$ for yaw DCT0 at matched
    noise levels. Error bars are 95\% confidence intervals; the gap is
    approximately zero early in the trajectory and increases to 0.955 at the
    final measured step.
    }
    \label{fig:app_amplification_gap}
\end{figure}

\cref{fig:app_amplification_gap} reports the same comparison as an
absolute difference with uncertainty. The measured gap is essentially zero
at steps 0--9, becomes positive at step 14, and increases to 0.955 at
step 19.

\subsubsection{Closed-Loop Alignment}
\label{app:lasi_online}

The closed-loop analysis contains 300 traces spanning 30 tasks and 16,707
valid replanning steps. Let $z_{e,t,m}$ and $y_{e,t,m}$ denote the source and
executed-action DCT coefficients for episode $e$, replanning step $t$, and
low-frequency camera mode
$m\in\mathcal M=\{\mathrm{P0,P1,P2,Y0,Y1,Y2}\}$.
To remove episode-level offsets, let $\bar z_{e,m}$ and $\bar y_{e,m}$ denote
the corresponding within-episode means and define
\begin{equation}
\operatorname{Align}(z,y)
=
\frac{1}{|\mathcal M|}
\sum_{m\in\mathcal M}
\operatorname{Corr}_{e,t}
\left(
z_{e,t,m}-\bar z_{e,m},
y_{e,t,m}-\bar y_{e,m}
\right).
\end{equation}

To test whether the action source remains aligned with later behavior rather
than merely reflecting a generic trajectory statistic, we circularly shift
the source schedule relative to the rollout. For the 75-position replan
schedule, define
\begin{equation}
    S(\delta)
    =
    \operatorname{Align}
    \left(
      \{z_{e,(t+\delta)\bmod 75,m}\},
      \{y_{e,t,m}\}
    \right).
\end{equation}

The unshifted statistic is 0.33763 and ranks first among all 75 circular
shifts. As a separate permutation control, we randomly permute the source
schedule $B=999$ times. No permuted statistic exceeds the observed value, and
the maximum permutation-null statistic is 0.06466. The corresponding
finite-sample permutation value is
\begin{equation}
p
=
\frac{
1+\sum_{b=1}^{B}
\mathbf 1[S_b^{\mathrm{perm}}\ge S_{\mathrm{obs}}]
}{
B+1
}
=
0.001.
\end{equation}

For the 197 complete 75-replan episodes, we additionally average each action
DCT coefficient across episodes at every replan position and correlate the
resulting mean trajectory with the explicit source schedule. The correlations
are $0.800,0.738,0.727$ for pitch DCT0--2 and
$0.803,0.808,0.639$ for yaw DCT0--2.

\begin{figure*}[t]
    \centering
    \includegraphics[width=0.78\textwidth]
    {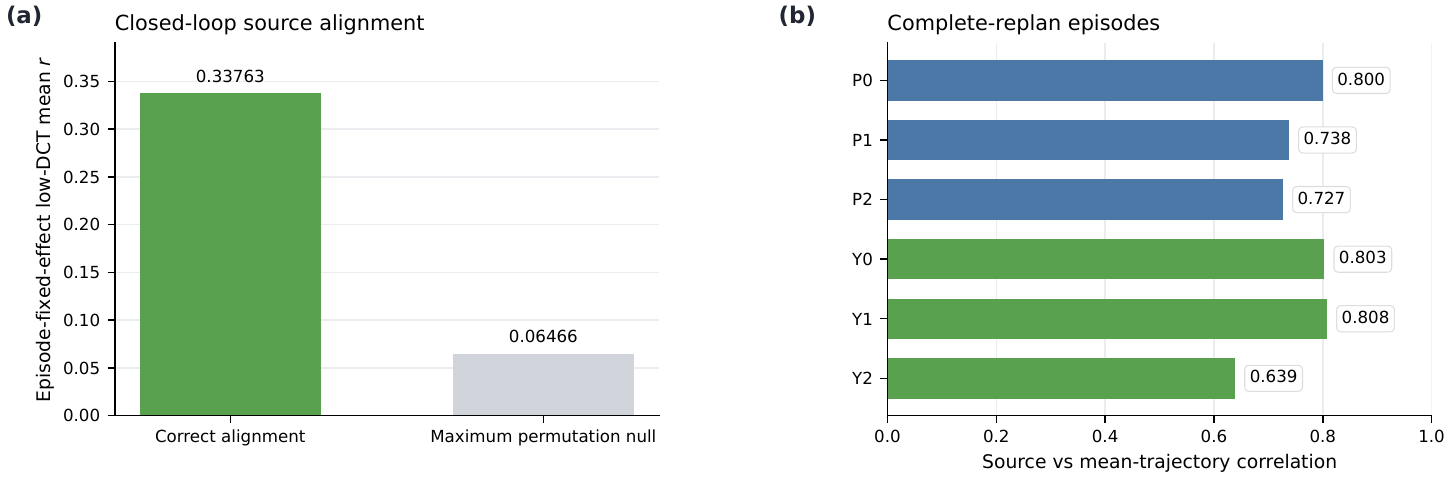}
    \caption{
    \textbf{Closed-loop LASI alignment.}
    \textbf{(a)} Episode-adjusted low-frequency association at the correct
    source/replan alignment compared with the maximum permutation-null
    statistic; the correct alignment additionally ranks first among all
    75 circular shifts.
    \textbf{(b)} Correlation between the explicit source schedule and the
    cross-episode mean trajectory for complete 75-replan episodes, shown for
    pitch and yaw DCT modes 0--2.
    }
    \label{fig:app_lasi_closed_loop}
\end{figure*}

\subsubsection{Closed-Loop Accumulation and Source Resampling}

The practical failure motivating LASI was most apparent when a single sampled
action source was reused across successive replanning steps. Closed-loop
behavior was strongly dependent on the reused source realization: for some
sources, a persistent directional bias accumulated into repeated in-place
rotation, and in severe cases almost no tasks were completed. Resampling the
action source between replanning steps largely removed this episode-level
failure pattern.

To characterize why repeated-source reuse can produce such behavior, for a
coarse camera quantity $\Delta_c$ at replanning step $c$, consider the
decomposition
\begin{equation}
    \Delta_c
    =
    \mu(\mathcal C_c)
    +
    b(Z_c)
    +
    \varepsilon_c,
\end{equation}
where $\mu(\mathcal C_c)$ collects condition-dependent behavior and
$b(Z_c)$ is a source-dependent low-frequency bias. This decomposition is
explanatory rather than a claim that the measured trajectory is fully
described by these terms.

If one source $Z$ is reused for $K$ cycles, the coherent component contributes
\begin{equation}
    \sum_{c=1}^{K}b(Z)
    =
    K\,b(Z),
\end{equation}
so a turning bias can be repeatedly reinforced. By contrast, if sources are
resampled independently and $b(Z_c)$ is approximately centered,
\begin{equation}
    Z_c\overset{\mathrm{i.i.d.}}{\sim}\mathcal N(0,I),
    \qquad
    \mathbb E
    \left[
      \sum_{c=1}^{K}b(Z_c)
    \right]
    \approx0,
\end{equation}
and
\begin{equation}
    \operatorname{Var}
    \left[
      \sum_{c=1}^{K}b(Z_c)
    \right]
    =
    K\operatorname{Var}[b(Z_c)].
\end{equation}

Thus, source resampling removes the coherent episode-level reuse of a single
source bias even though stochastic action variation remains. Consistent with
this analysis, our closed-loop evaluation resamples the action source at every
replanning step. We treat this as an evaluation-time mitigation of coherent
accumulation rather than a mechanism-level correction of the underlying LASI
sensitivity.

\section{Scope, Limitations, and Interpretation Boundaries}
\label{app:limitations}

\subsection{Method and Evaluation Scope}

GameWAM is designed as a low-level closed-loop controller rather than a
standalone high-level planner. Its policy is conditioned on current visual,
language, historical, and proprioceptive context, but it does not explicitly
maintain a symbolic task graph, inventory plan, recipe representation, or
long-horizon search process. Failures that require semantic decomposition,
exploration, or multi-stage resource planning therefore remain outside the
intended scope of the block-cycle controller. A natural extension is to pair
GameWAM with a higher-level planner while retaining GameWAM as the native
keyboard--mouse policy.

This distinction is particularly relevant to GUI interaction. Crafting,
smelting, and inventory manipulation combine low-level cursor and keyboard
control with structured procedural knowledge. Higher-level agents can
externalize such procedures through symbolic actions, explicit plans, or
dedicated executors, whereas GameWAM must realize them through the same
low-level native-action policy. The scripted GUI data provides procedural
supervision for these interactions, but the current model does not separately
represent recipes or explicitly plan multi-step GUI procedures. Future systems
could combine GameWAM with structured procedural representations or a
higher-level planner while preserving closed-loop visual feedback at the
execution level.

The current persistent history is also predominantly visual. Its auxiliary
objective encourages the compressed cross-cycle representation to remain
predictive of the current clean visual feature, but it does not directly
supervise which historical information should be retrieved or used for
downstream action generation and task completion. Future work could introduce
task- or action-aware history objectives, explicit retrieval supervision,
cross-cycle credit assignment, or multimodal memory that jointly represents
visual observations, actions, inventory state, and higher-level task context.

Our empirical evaluation is restricted to digital game environments. Minecraft
and ViZDoom provide diverse visual dynamics and native action spaces, but they
do not establish transfer to physical control, where sensing delay, actuation
noise, safety constraints, and contact dynamics differ substantially. The
Minecraft results also depend on the available VPT/MineStudio data mixture; the
current experiments do not isolate every possible dataset-composition factor. More broadly, current
Minecraft Universe (MCU) training and evaluation settings largely focus on
atomic task completion and individual interaction events. While useful for
measuring specific capabilities, they provide limited supervision and
evaluation for temporally extended behaviors where intermediate decisions
influence future states and actions. Constructing high-quality chain-of-
interaction trajectories with coherent intent, evolving state context, and
aligned native actions remains challenging, and represents an important
direction for future game-agent datasets and evaluations.

\subsection{Interpretation of the LASI Diagnostics}
\label{app:claim_boundaries}

The combined LASI evidence supports four specific statements. First, an
explicit intervention on low-frequency source coefficients causally changes
the matching low-frequency action output under fixed conditioning. Second, the
effect is substantially stronger at low than at high temporal frequencies.
Third, source transfer is already measurable in a single analytic forward
pass. Fourth, the iterative model path amplifies the effect later along the
generative trajectory, and a statistically aligned component remains
detectable in closed-loop trajectories.

These results remain diagnostic rather than a complete mechanistic account.
They do not identify a unique internal layer or training-time pathway, imply
that conditioning information is absent, or establish that source variation
explains the complete closed-loop trajectory. Empirically, the clearest LASI
effect appears in continuous camera control. A plausible contributor is the
pronounced low-frequency structure of camera trajectories in the training
data: meaningful camera motion often evolves smoothly over short horizons,
potentially making low-frequency source structure especially easy to align
with and propagate through the generated action trajectory. This provides a
possible data-level explanation for the concentration of the observed effect
in camera motion.

Resampling the action source across replanning steps breaks repeated exposure
to the same source bias and largely removes its harmful coherent accumulation
in closed-loop interaction, but it does not remove the underlying source
sensitivity. We therefore also explored training-time interventions aimed at
the formation of LASI itself. These included frequency-selective objectives
that directly discourage low-frequency source--action coupling, consistency
objectives that reduce dependence on particular source realizations under the
same conditioning, and additional supervision that strengthens the dependence
of generated actions on conditioning. Across these directions, stronger
suppression could interfere with normal action learning or degrade closed-loop
performance, whereas other variants preserved task behavior but left
substantial low-frequency source dependence. None therefore provided a robust
mechanism-level correction without an accompanying trade-off.

We consequently use source resampling as an effective deployment-time
mitigation while treating mechanism-level removal of LASI as an open problem.
A desirable solution should prevent stochastic source structure from becoming
a persistent low-frequency control bias without suppressing legitimate smooth
motion, condition-compatible action diversity, or closed-loop responsiveness.
Developing such objectives remains an important direction for generative
action policies.

\end{document}